\documentclass{article}

\usepackage{microtype}
\usepackage{graphicx}
\usepackage{subfigure}
\usepackage{booktabs} 
\usepackage{multirow}

\usepackage{hyperref}

\usepackage[accepted]{icml2025}

\usepackage{amsmath}
\usepackage{amssymb}
\usepackage{mathtools}
\usepackage{amsthm}

\usepackage[capitalize,noabbrev]{cleveref}

\theoremstyle{plain}

\theoremstyle{definition}

\theoremstyle{remark}

\usepackage[textsize=tiny]{todonotes}

\icmltitlerunning{Mind Your Step (by Step)}

\begin{document}

\twocolumn[
\icmltitle{Mind Your Step (by Step): Chain-of-Thought can Reduce Performance \\ on Tasks where Thinking Makes Humans Worse}



\icmlsetsymbol{equal}{*}

\begin{icmlauthorlist}
\icmlauthor{Ryan Liu}{equal,1}
\icmlauthor{Jiayi Geng}{equal,1}
\icmlauthor{Addison J. Wu}{1}
\icmlauthor{Ilia Sucholutsky}{2}
\icmlauthor{Tania Lombrozo}{3}
\icmlauthor{Thomas L. Griffiths}{1,3}
\end{icmlauthorlist}

\icmlaffiliation{1}{Department of Computer Science, Princeton University, Princeton, NJ, USA}
\icmlaffiliation{2}{NYU Center for Data Science, New York University, New York, NY, USA}
\icmlaffiliation{3}{Department of Psychology, Princeton University, Princeton, NJ, USA}

\icmlcorrespondingauthor{Ryan Liu}{ryanliu@princeton.edu}
\icmlcorrespondingauthor{Jiayi Geng}{jiayig@princeton.edu}

\icmlkeywords{Machine Learning, ICML}

\vskip 0.15in
]



\printAffiliationsAndNotice{\icmlEqualContribution} 

\begin{abstract}

Chain-of-thought (CoT) prompting has become a widely used strategy for improving large language and multimodal model performance. 
However, it is still an open question under which settings CoT systematically {\em reduces} performance. In this paper, we seek to identify the characteristics of tasks where CoT reduces performance by drawing inspiration from cognitive psychology, focusing on six representative tasks from the psychological literature where deliberation hurts performance in humans. In three of these tasks, state-of-the-art models exhibit significant performance drop-offs with CoT (up to 36.3\% absolute accuracy for OpenAI o1-preview compared to GPT-4o), while in others, CoT effects are mixed, with positive, neutral, and negative changes. While models and humans do not exhibit perfectly parallel cognitive processes, considering cases where thinking has negative consequences for humans helps identify settings where it negatively impacts models. By connecting the literature on human verbal thinking and deliberation with evaluations of CoT, we offer a perspective for understanding the impact of inference-time reasoning.
\end{abstract}

\section{Introduction}
\label{sec:introduction}

\begin{figure*}[h]
\centering
    \includegraphics[width=0.7\textwidth]{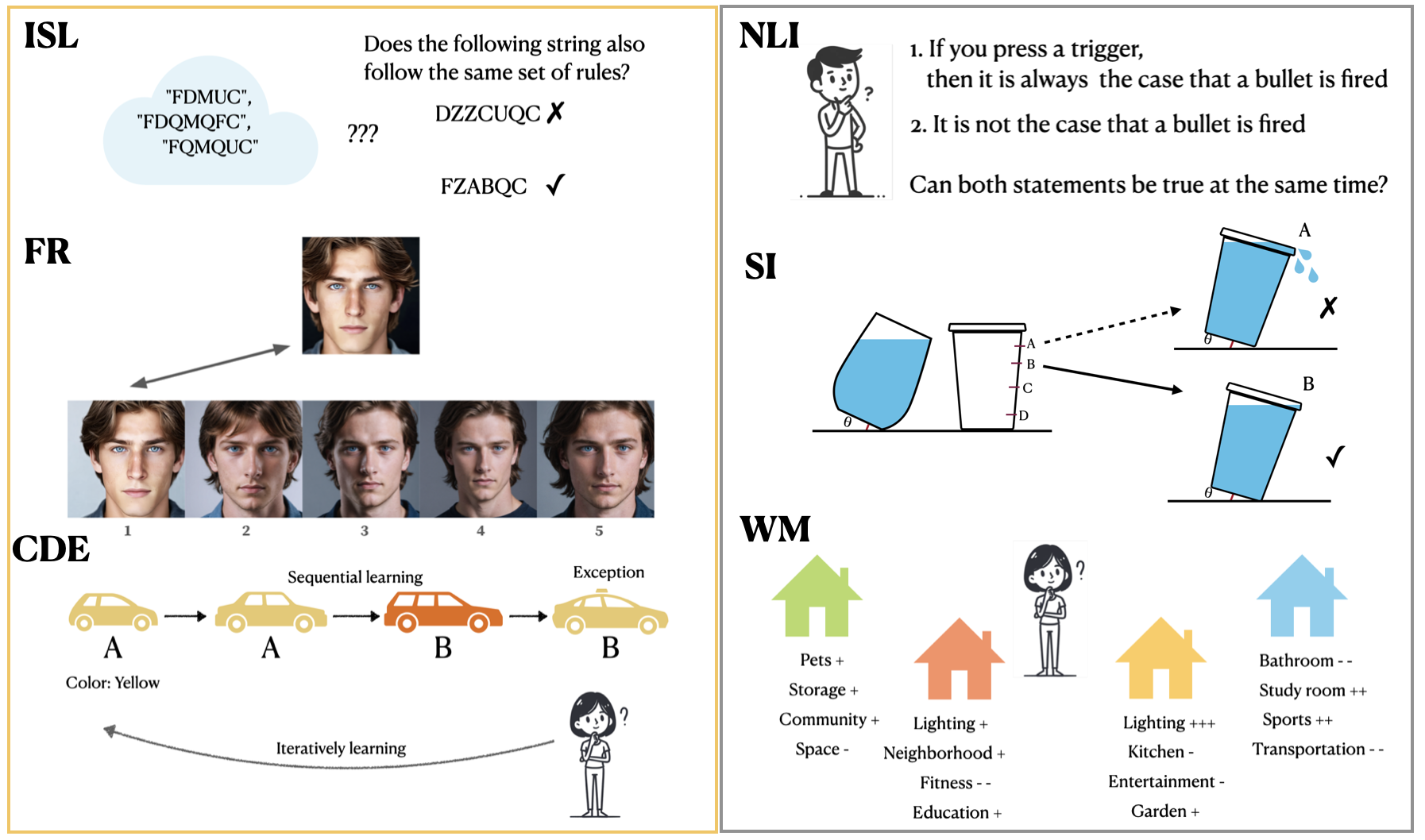}
    \caption{Tasks evaluated for reductions in performance from CoT. \textbf{Left:} Implicit Statistical Learning (ISL): Classifying if strings were generated by an artificial grammar. Face Recognition (FR): Recognition of a face from a set that shares similar descriptions. Classifying Data with Exceptions (CDE): Learning labels in the presence of exceptions. \textbf{Right:} Natural Language Inference (NLI): Recognizing a logical inconsistency. Spatial intuitions (SI): Tilting water glasses. Working Memory (WM): Aggregating features for a decision. Humans show reductions in performance when engaging in verbal thinking in all tasks, and LLMs and LMMs show similar effects in the first three.}
    \label{fig:main_fig}
\end{figure*}

Chain-of-thought~\citep{wei2022chain, nye2021show} is a widely used technique for prompting large language and multimodal models (LLMs and LMMs), instructing models to ``think step-by-step'' or providing other structure that should be incorporated into the response. Large meta-studies have shown that this technique improves the performance of models in many tasks, particularly those involving symbolic reasoning~\citep{sprague2024cot}. More generally, inference-time reasoning has become a default component of the newest LLMs and LMMs such as OpenAI's o-series models~\citep{openai2024o1} and Claude's web interface and thinking models~\citep{anthropic_2025_thinking}. However, there also exist cases where CoT {\em decreases} performance, but the literature has not identified any patterns as to when this happens. As inference-time reasoning becomes the new norm in deployed frontier models, it is imperative to understand and predict when CoT has a negative effect on model performance. 

A key challenge for determining the limits of CoT is the sheer variety of tasks for which LLMs and LMMs are used. While the machine learning community has dedicated great efforts towards developing a large set of benchmarks for these models~\citep[e.g.,][]{hendrycks2020measuring, suzgun2022challenging}, applications of models extend beyond benchmarks to diverse contexts and variations of tasks that could all potentially affect performance. Exploring this vast space to identify settings where CoT has negative effects is a daunting problem. This motivates the need to develop heuristics to help us identify risky cases that could pose challenges for inference-time reasoning.

In this paper, we explore the effectiveness of a heuristic formed by drawing a parallel between CoT reasoning and humans engaging in verbal thought~\citep{lombrozo2024learning}. Specifically, we explore whether tasks for which thinking decreases human performance are also tasks where CoT harms model performance.
This heuristic is motivated by the idea that in some cases, the tasks themselves, in conjunction with shared traits between humans and models, result in verbal thinking negatively impacting performance. 
However, models and humans also have different capabilities and consequently different constraints for some tasks~\citep{griffiths2020understanding,shiffrin2023probing,mccoy2023embers}. For example, LLMs have context lengths that exceed human memory limitations. Thus, we do not expect our heuristic to predict model performance perfectly, but rather allow us to quickly identify cases where CoT has a significant negative impact.

To explore our approach, we first draw on the psychology literature for a collection of tasks where verbal thinking hurts human performance~\citep[\textit{inter alia}]{schooler1990verbal, dijksterhuis2004think, van2008intentional}. We then organized these into six prominent task archetypes, and selected the most representative study from each (see Figure \ref{fig:main_fig}). We adapted these tasks to properly evaluate LLMs and LMMs and measured performance with and without CoT. In our experiments, we found large performance decreases from CoT in three archetypes: tasks that involve implicit statistical learning, tasks where language is ill-suited to represent stimuli, and tasks that involve learning labels that contain exceptions to generalizable rules. For the three other archetypes, we did not see consistent decreases in performance from CoT. For these, we suggest explanations for why CoT does not decrease performance based on meaningful differences between humans and models. 

In representative tasks for each of the first three archetypes, CoT drastically decreased performance across models. First, in implicit statistical learning, we observed a drop of 36.3\% in the absolute performance of o1-preview compared to directly prompting GPT-4o, along with consistent reductions in accuracy from CoT across eight other state-of-the-art LLMs. Next, for a task involving visual stimuli ill-represented by language, we found that CoT reduced performance in all six vision-language models tested. And third, when learning labels that contain exceptions to generalizable rules, CoT increased the number of iterations it took all models to learn the correct labels, by up to 331\% more.

In contrast, for the latter three archetypes, we did not observe consistently negative effects caused by CoT, but instead a mixed collection including positive and neutral instances. For these tasks, we suggest plausible explanations for why models do not perform worse while humans do. 
When LLMs had access to relevant priors that humans lacked, such as in the logical inconsistency task where participants were not adept in formal logic, performance generally improved using CoT as LLMs could tap into existing logical frameworks while reasoning whereas human participants could not. 
Conversely, when models lacked access to relevant priors, such as in a task where motor simulation was responsible for improved performance (relative to verbal thinking) in humans, performance was roughly equal between conditions. Finally, having access to long context windows compared to human working memory synergized with CoT to improve model performance in a preference aggregation task involving a myriad of features. 
These cases highlight the importance of understanding differences between humans and models when translating psychological results to predictions about model performance. 

To evaluate whether to use CoT on unseen tasks, one could search for stimulus patterns (such as reliance on another modality) that are based in psychological findings, and predict that e.g., due to verbal overshadowing, performance using CoT would also be poor.
While our work does not create a definitive classifier for when CoT should not be used, our results provide valuable scientific observations supporting a parallel between human deliberation and model CoT failures. 
These failure cases we derive are uniquely informative for studying the limits of CoT because of existing psychology literature that explains why these failures happen (see Section~\ref{sec:methods}). 
Such parallels also suggest that these failures we observe are not arbitrary, but rather reflect deeper patterns in reasoning.

The remainder of the paper is structured as follows: We cover related work surrounding CoT and intersections between AI models and psychology in Section~\ref{sec:related_work}. We ground our approach in the psychology literature and identify six archetypes of tasks for which thinking reduces human performance in Section~\ref{sec:methods}. In Section~\ref{sec:experiments}, we describe the six representative tasks corresponding to each archetype, how we implemented these tasks as model evaluations, and summarize our findings. We then contextualize our findings and discuss limitations in our work in Section~\ref{sec:discussion}.

\section{Related work}
\label{sec:related_work}

\subsection{Inference-time reasoning}

Chain-of-thought prompting aims to improve the performance of language-based models by encouraging them to generate an intervening string of tokens that increases the probability of producing the correct answer \citep{wei2022chain,nye2021show}. This approach can result in significant performance improvements in language \citep{zhang2022automatic} and vision \citep{zhang2023multimodal} tasks, hypothesized to be a consequence of exploiting local structure in language \citep{prystawski2024think}. However, a recent metastudy suggests that gains from using CoT are primarily in mathematical and symbolic reasoning tasks, and other areas such as text classification can instead see decreases in performance~\citep{sprague2024cot}. Yet, even the reasoning capabilities on symbolic tasks are also fragile~\citep{mirzadeh2024gsm}. In the literature, there are no fine-grained patterns that explain under which cases CoT performs poorly. 
In settings such as planning, studies have shown little benefit from using CoT \citep{kambhampati2024llms}. Furthermore, CoT can also increase the frequency of harmful outputs~\citep{shaikh2023second}. Despite these results, the default expectation is that CoT improves performance. For example, a recent update to a popular benchmark cited the fact that CoT results in an improvement on the new benchmark but decreased performance on the original as an indicator that the new benchmark is better \citep{wang2024mmlu}. This expectation has likely driven the tendency towards a default use of CoT and inference time reasoning in the latest models.

\subsection{Psychological methods as a tool for studying models}

Since the introduction of LLMs, there has been growing interest in understanding the connections between models and human minds \citep{hardy2023large}. 
Human cognition is often studied using well-controlled tasks involving carefully curated datasets designed to test specific hypotheses. The availability of these datasets, and the fact that they often consist mainly of text and/or images, have led to these tasks from the psychology literature to quickly become popular methods for evaluating and understanding LLMs and LMMs \citep[e.g.,][]{binz2023using,coda2024cogbench}. 
Recent studies that leverage insights or datasets from psychology have evaluated the representational capacity of LLMs \citep{frank2023baby}, explored how RLHF and CoT lead to different outcomes when trying to make models helpful and honest \citep{liu2024largeb}, and
compared human and machine representations via similarity judgments \citep{peterson2018evaluating,marjieh2023words, marjieh2023language, marjieh2024large}. Studies have also found that LLMs over-estimate human rationality \citep{liu2024largea}, identified incoherence in LLM probability judgments \citep{zhu2024incoherent}, identified susceptibility to linguistic illusions in LLMs \citep{marjieh2024rational}, and uncovered LLMs' underlying social biases \citep{bai2024measuring}. Other works have used storytelling to understand episodic memory in LLMs \citep{cornell2023role}, constructed prompts using theories of metaphor \citep{prystawski2022psychologically}, discovered cross-linguistic variability in LLM representations \citep{niedermann2023studying}, and probed the roles of language and vision for in-context learning in VLMs \citep{chen2024analyzing}. 
Many of these studies start with a phenomenon in human cognition and explore whether there is an analog to it in AI models. Our work follows this approach by associating the well-studied impact of deliberation on human performance with the effects of CoT on model performance.  

\section{Approach: When thinking impairs humans}
\label{sec:methods}

A large body of psychological research has investigated effects of verbal thinking (often explicit ``deliberation'') on memory, learning, judgment, and decision-making. Very often these effects are positive. For example, people who spend more time deliberating are
more likely to respond correctly on questions that initially trigger an intuitive but incorrect response~\citep{travers2016time}. However, there are also cases in which verbal thinking can impair performance, often involving a mismatch between the representations or types of processing that best support the task and those induced by verbal thinking~\citep{schooler2002re}. 

A canonical setting for such effects is implicit statistical learning. Examples include artificial grammar learning studies, where participants are presented with sequences of letters or phonemes that conform to some structure (e.g., a finite state grammar) and are asked to recognize well-formed sequences. Studies found that participants can differentiate well-formed sequences from those that are not, but cannot verbalize the basis for their judgments \citep{aslin2012statistical, romberg2010statistical}. Some (but not all) studies further find that receiving explicit instructions to identify rules in verbal form impairs performance \citep{reber1976implicit}. In this task archetype, the implicit statistical structure of learned patterns are often better recognized without verbal descriptions. We use the original artificial grammar learning task shared by \citet{reber1977implicit}, \citet{whittlesea1993incidentally}, and \citet{van2008intentional} to test models.

Another class of cases concerns a phenomenon termed verbal overshadowing. In a classic demonstration, instructions to verbalize a face led to impaired facial recognition compared to a condition in which participants did not~\citep{schooler1990verbal}. Such effects have been found for other perceptual stimuli \citep{fiore2002did, melcher1996misremembrance}, but do not extend to stimuli that are easy to verbalize (such as a spoken statement) \citep{schooler1990verbal} or to logical problem solving \citep{schooler1993thoughts}. This task archetype features stimuli that are better processed in another modality, and consequently humans perform worse when they are forced to solve these tasks verbally. We adapt the seminal facial recognition demonstration from \citet{schooler1990verbal} to test multimodal models. 

As a third category, studies find that asking people to generate verbal explanations for their observations supports the discovery of broad and simple patterns \citep{edwards2019explanation, walker2017explaining, williams2010role, williams2013explanation}. But when the stimuli are designed such that these broad and simple patterns contain exceptions, participants who were prompted to explain learned more slowly and made more errors \citep{williams2013hazards}. 
These effects are thought to arise from a mismatch between the representations or processes induced by a form of thinking (in this case, explaining) and those which best support task performance \citep{lombrozo2016explanatory}. To test models, we adapt experiments from \citet{williams2013explanation}---the landmark study that demonstrated these failure effects.

The effects reviewed so far plausibly concern impairments that arise from the representational limitations of language and the generalization of patterns found in language: language is not well-suited to encoding fine-grained perceptual discriminations (e.g., in face recognition), and language readily encodes some kinds of relationships (such as deductive entailment, or simple and broad patterns) but is less well-suited or employed for others (such as complex grammars, or patterns with arbitrary exceptions). Given that LLMs are likely to share limitations that arise from language and generalization, we might expect LLMs to exhibit patterns of impairment that mirror those found for humans on these tasks. We test these predictions using the selected study paradigms in Section~\ref{sec:experiments}.

Prior work has documented additional impairments in humans from verbal thinking, but for some it is less clear if they should generalize to LLMs. For example, explaining how inconsistent statements could be true makes human participants less likely to recognize a logical inconsistency \citep{khemlani2012hidden}. 
However, these participants are typically untrained in formal logic, whereas LLMs are familiar with logical manipulation via their vast training corpuses. 
Prior work has also found that verbal thinking can be less accurate than visual or motor simulation (\citealt{schwartz1999inferences}; see also \citealt{aronowitz2020learning, lombrozo2019learning}), but this is a consequence of information encoded in visual and motor representations that are likely not available to models. Finally, humans sometimes make poor choices when they deliberate over multi-dimensional aggregation problems with numerous features \citep{dijksterhuis2004think} -- plausibly a consequence of memory limitations that are not faced by LLMs. We anticipate that for tasks like these, CoT is less likely to reduce performance. 
We adapt one task from each of these archetypes (explanations, motor simulation, memory) and also apply them to test LLMs.

\section{Experiments}
\label{sec:experiments}

For each of the six representative studies on human verbal thinking described in Section~\ref{sec:methods}, we test the performance of a collection of LLMs or LMMs with and without CoT. 
For each task, we massively scale up the study stimuli and adapt the task to appropriately test these models.
We release the six scaled-up task datasets as a human overthinking benchmark that ML practitioners can use, available at \url{https://github.com/JiayiGeng/CoT_overthinking}.

\subsection{Implicit statistical learning}
\paragraph{Task.} The first class of tasks we examine are those involving implicit statistical learning. As described in Section~\ref{sec:methods}, psychology studies found that data that contain statistical patterns can sometimes be better generalized by humans when those patterns are not linguistically described. We explore whether this also holds for LLMs by replicating the task of artificial grammar learning~\citep{reber1977implicit, whittlesea1993incidentally, van2008intentional}. In this task, artificial ``words'' are constructed using finite-state grammars (FSGs) and participants are tasked with identifying which words belong to the same pattern (are generated by the same FSG). 

To test this, we constructed 4400 classification problems corresponding to 100 randomly sampled unique FSGs that were structurally similar to those in \citet{fallshore1993post}. Each classification problem consisted of 15 training examples generated from the grammar, and the model was given a new example and asked to classify it. For each FSG, models were asked to classify 44 words, where 22 words belonged to the FSG and 22 did not. Words not belonging to the grammar were generated by replacing one letter of an existing word in the grammar. Details on problem generation are provided in Appendix~\ref{app:agl_generation}.

To test whether our results generalize across the broader task archetype, we also examined the effects of CoT on iteratively simpler versions of the problem; we reduced the complexity of our 6-node FSG to \{5, 4, 3\} nodes via edge contraction (see Appendix~\ref{app:agl_complex}), creating a series of distinct implicit statistical learning tasks of varying difficulty.

\paragraph{Human failure.} In the artificial grammar learning task, humans prompted to verbalize performed more poorly than those who were not told to do so~\citep{fallshore1993post}. 
We investigate whether CoT will similarly reduce LLM performance.

\paragraph{Models and prompts.} We test several open- and closed-source models: OpenAI o1-preview, GPT-4o, Claude 3.5 Sonnet, Claude 3 Opus, Gemini 1.5 Pro, Llama 3.1 70B \& 8B Instruct, and Llama 3 70B \& 8B Instruct. Due to cost limitations, we test o1-preview on a subset. We could not remove o1's inference-time reasoning, and thus used GPT-4o as its zero-shot comparison. We mainly evaluated zero-shot and CoT prompts (Appendix~\ref{app:agl_prompts}), but also experimented with tree-of-thought~\citep[Appendix~\ref{app:agl_tot}]{yao2023treethoughtsdeliberateproblem}, robustness to sampling temperature (Appendix~\ref{app:agl_temp}), and in-context steering towards reasoning methods more suited for the statistical structure of the task (Appendix~\ref{app:agl_steering}).

\paragraph{Results.} We find large reductions in performance when using CoT compared to zero-shot prompting, as shown in Table~\ref{tab:agl_results}. When run on a randomly selected subset of 440 problems, o1-preview, with inference-time reasoning built into its responses, has a 36.3\% absolute accuracy decrease compared to GPT-4o zero-shot on the same subset. Similarly, while there is limited performance change between conditions for Claude 3.5 Sonnet, both conditions perform worse than Claude 3 Opus zero-shot. Across other models, we find consistent decreases in performance with CoT: 23.1\% in GPT-4o, 8.00\% in Claude 3 Opus, 6.05\% in Gemini, and 8.80\% in Llama 3.1 70B Instruct. Weaker models such as Llama 3.1 8B Instruct have smaller drops in absolute performance, but these are still substantial when contextualized against random chance (50\%). As such, the differences caused by CoT remain statistically significant. 

Additionally, CoT reductions in performance extend across task complexity (Appendix~\ref{app:agl_complex}), and while both tree-of-thought and in-context reasoning steering helps performance, neither method meaningfully bridges the gap towards zero-shot performance (Appendices~\ref{app:agl_tot}, \ref{app:agl_steering}). 

\begin{table}
\centering
\caption{For artificial grammar learning, CoT greatly reduces performance in a majority of models. Random performance is 50\%. 
}
\label{tab:agl_results}
\resizebox{.45\textwidth}{!}{
\begin{tabular}{lcccc}
\hline
                  & Zero-shot & CoT     & \begin{tabular}[c]{@{}c@{}}decrease\\ (absolute)\end{tabular} & \multicolumn{1}{c}{$p$-value~~~~}                  \\ \hline
GPT-4o (subset)   & 94.00\%   & -       & \multirow{2}{*}{$36.30\%$}                                       & \multirow{2}{*}{$<0.0001$} \\
o1-preview (subset)   & -         & 57.70\% &                                                                &                          \\
GPT-4o            & 87.50\%   & 64.40\% & 23.10\%                                                        & $<0.0001$                 \\
Claude 3 Opus     & 70.70\%   & 62.70\% & 8.00\%                                                         & $<0.0001$                 \\
Claude 3.5 Sonnet & 65.90\%         & 67.70\% & -1.80\%                                                         & \hspace{5.5pt}$0.969$                  \\
Gemini 1.5 Pro    & 68.00\%   & 61.95\% & 6.05\%                                                         & $<0.0001$                 \\
Llama 3 8B Instruct       & 59.70\%   & 57.90\% & 1.80\%                                                         & $<0.05$                  \\
Llama 3 70B Instruct       & 60.50\%   & 58.30\% & 2.20\%                                                         & $<0.05$                  \\
Llama 3.1 8B Instruct      & 53.52\%   & 51.54\% & 1.98\%                                                         & $<0.0001$                  \\
Llama 3.1 70B Instruct     & 65.90\%   & 57.10\% & 8.80\%                                                         & $<0.0001$                 \\ \hline
\end{tabular}
}
\end{table}

\subsection{Facial recognition}
\paragraph{Task.} The second archetype of tasks we identify from the literature involves verbal overshadowing. We study this case using a classic face recognition task, where participants are first shown a face and then asked to select an image of the same person from a set of candidates~\citep{schooler1990verbal}. While psychological studies often include a distractor task between the initial face and the candidates to increase the difficulty, we did not use these for LMMs due to their weak performance. We scale this task from one recognition problem to a novel synthetic dataset of 500 problems across 2500 unique faces. For each problem, all faces were given the same described attributes for seven features: race, gender, age group, eye color, hair length, hair color, and hair type. We then generated a pair of images of the same person and four images of other people matching this description using stable-image-ultra~\citep{stabilityai_api_2024}, which we selected based on generation quality. We adjusted the generation process to ensure that the pair clearly consisted of the same person, while the others clearly did not (see Appendix~\ref{app:face_generation} for further details). One of the pair was selected to be the initial stimulus, while the other was shuffled with the four images to create the set of candidate answers. 
Models were asked to identify which candidate matched the person from the initial stimulus. 

To test whether our results generalize across the broader archetype, we also tested a less difficult case where the images' descriptions within the answer options were not required to be the same (see Appendix~\ref{app:face_difficult}). 

\paragraph{Human failure.} Participants prompted to verbally describe the faces performed worse than those who were not~\citep{schooler1990verbal}.
We investigate if CoT similarly reduces performance on the task in LMMs.

\paragraph{Models and prompts.} We evaluated this task on several open- and closed-source LMMs: GPT-4o, Claude 3.5 Sonnet, Claude 3 Opus, Gemini 1.5 Pro, InternVL2 26B, and InternVL2 Llama3 76B. Other multimodal models were not considered as they did not support multiple image input. We mainly evaluated zero-shot and CoT prompts (available in Appendix~\ref{app:face_prompts}), but also experimented with a binary classification prompt (i.e., ``is this the same person?'') to rule-out explanations of CoT disrupting image ordering (Appendix~\ref{app:face_order}). 

\paragraph{Results.} We find that all six LMMs tested show a drop in performance when performing CoT (see Table~\ref{tab:vision_results}). Weaker models often answered that ``all images are of the same person'', resulting in accuracies below the random chance rate of 20\%. However, even in these cases we observe decreases in performance due to CoT. We find similar reductions in the reduced difficulty and binary classification settings, which 1) demonstrates generalizability across the task archetype; and 2) suggests that CoT affects the reasoning process, rather than causing surface-level ordering failures. 

\begin{table}
\centering
\caption{CoT reduces performance for facial recognition across all six LMMs. Random performance is 20\%.}
\label{tab:vision_results}
\resizebox{.48\textwidth}{!}{%
\begin{tabular}{@{}lccccc@{}}
\toprule
                     & Zero-shot & CoT     & \begin{tabular}[c]{@{}c@{}}decrease\\ (absolute)\end{tabular} & \begin{tabular}[c]{@{}c@{}}decrease\\ (relative)\end{tabular} & $p$-value         \\ \midrule
GPT-4o               & 64.00\%   & 51.20\% & 12.80\%                                                                   & 20.00\%                                                                   & $<0.01$\\
Claude 3 Opus        & 44.00\%   & 29.60\% & 14.40\%                                                                   & 32.73\%                                                                   &   $<0.0001$               \\
Claude 3.5 Sonnet    & 97.80\%   & 94.80\% & 3.00\%                                                                    & 3.07\%                                                                    &      $<0.05 $        \\
Gemini 1.5 Pro       & 66.00\%   & 54.60\% & 11.40\%                                                                   & 17.27\%                                                                   &      $<0.05$         \\
InternVL2 26B        & 9.20\%    & 6.00\%  & 3.20\%                                                                    & 34.78\%                                                                   &      $<0.05$         \\
InternVL2 Llama3 76B & 15.77\%   & 13.77\% & 2.00\%                                                                    & 12.68\%                                                                   &      $0.44$        \\ \bottomrule
\end{tabular}%
}
\end{table}

\subsection{Classifying data with rules that contain exceptions}
\label{sec:experiments_vehicle}

\paragraph{Task.} A third class of tasks where CoT may harm performance is learning to classify exemplars when there are exceptions to generalizable rules. As mentioned in Section~\ref{sec:methods}, when humans try to explain the category membership of exemplars, they tend to hypothesize simple classification rules, which can lead to inefficient learning when data contain arbitrary exceptions to these rules. 

To study if this phenomenon extends to CoT, we replicate a multi-turn vehicle classification task from~\citet{williams2013hazards}, in which participants try to correctly assign binary labels to a list of vehicles. Participants are given feedback after each prediction, and conduct multiple passes over the list until they label all vehicles correctly in a single pass or exceed the maximum number of tries. 
Vehicles in the task contained one feature that was almost fully correlated (80\%) with the classification label, three features with no relation to the label, and one feature that individually identified the vehicle. Thus, participants could either try to learn a generalizable rule from the highly correlated feature but fail due to the exceptions, or they could learn the individual mappings from the identifying feature to the corresponding label. Human participants who were prompted to explain the classification of exemplars performed worse because they tended to attempt the former strategy. 

Participants in the treatment condition of the original study were prompted to deliberate after receiving feedback. To more explicitly include inference-time reasoning, we modify the point at which deliberation is prompted, instead asking the LLM to perform CoT before making each prediction. 
In total, we constructed $2400$ vehicles --- split into $240$ lists of ten vehicles each --- and measured LLMs' abilities to learn the labels of each list across up to $15$ passes (see Appendix \ref{app:vehicle_generation} for details). Memory was implemented by including previous problems, guesses, and feedback in context. 

To test the robustness of our results, we also tested an easier case where the feature without exceptions (license plate) is replaced with a binary oracle feature (see Appendix~\ref{app:vehicle_difficulty}).

\paragraph{Human failure.} When learning, people tended to reason about generalizable rules when conducting explanations, increasing the time needed to learn labels for the entire list~\citep{williams2013hazards}. 

\paragraph{Models and prompts.}\hspace{0pt}We evaluated this task on GPT-4o, Claude 3.5 Sonnet, and Claude 3 Opus. Other models such as Llama 3.1 70B Instruct were insufficient at multi-turn long context conversation, which made outputs unusable. We mainly used direct and CoT prompts (see Appendix~\ref{app:vehicle_prompts}), and also investigated the effects of a series of steering prompts with varied strength (Appendix~\ref{app:agl_steering}).

\begin{table}
\centering
\caption{Average number of rounds for models to learn labels greatly increases when using CoT. GPT-4o with CoT often never completely learned the labels (15 rounds). }
\label{tab:vehicle_results}
\resizebox{.48\textwidth}{!}{%
\begin{tabular}{@{}lccccl@{}}
\toprule
                  & Direct & CoT  & \begin{tabular}[c]{@{}c@{}}Rounds increase\\ (absolute)\end{tabular} & \begin{tabular}[c]{@{}c@{}}Rounds increase\\ (relative)\end{tabular} & \hspace{4pt}$p$-value   \\ \midrule
GPT-4o            & 2.9       & 12.5 & 9.6 & 331\%                                                                   &       $<0.0001$   \\
Claude 3.5 Sonnet & 2.3       & 6.4  & 4.1                                                                     & 178\%                                                                   & $<0.0001$ \\
Claude 3 Opus     & 2.4       & 5.5  & 3.1                                                                     & 129\%                                                                   &    $<0.05 $      \\ \bottomrule
\end{tabular}
}
\end{table}

\paragraph{Results.} We find that CoT drastically increases the number of passes for the model to learn all labels correctly. Averaged across the 240 lists, GPT-4o with CoT needed more than four times the number of passes to learn the labels compared to direct prompting, while Claude 3.5 Sonnet and 3 Opus both needed more than double (see Table~\ref{tab:vehicle_results}). 

\begin{figure}
    \centering
    \includegraphics[width=0.75\linewidth]{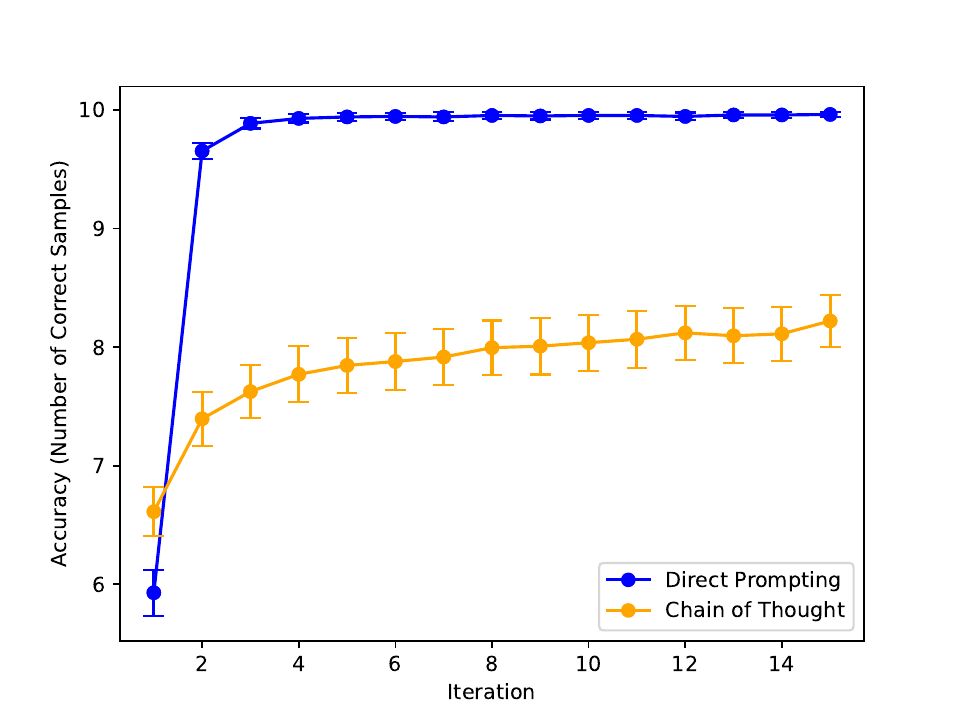}
    \caption{Number of correct objects classified per round for GPT-4o. CoT stagnates at 80\%, the proportion which are not exceptions.}
    \label{fig:learning_curve}
\end{figure}

We also investigated the per-round accuracy of GPT-4o and found that direct prompting resulted in the model attaining perfect classification on the second or third iteration, while with CoT the model was only able to correctly classify 8/10 objects on average after 15 iterations (see Figure~\ref{fig:learning_curve}). The model was unable to surpass this degree of accuracy, likely due to CoT biasing the model to rely on the seemingly generalizable rules from the exemplars, while down-weighing the usefulness of contextual tokens that explicitly contained all of the correct answers. 

We found similar reductions of performance from CoT even in the case with a binary oracle feature (Appendix~\ref{app:vehicle_difficulty}), but CoT performance reached zero-shot levels with strong steering prompts that explicitly told the model to match license plates (Appendix~\ref{app:vehicle_steering}). This demonstrates that CoT reasoning is indeed able to match zero-shot performance, but its default does not match the optimal reasoning space. 

\subsection{Tasks with a mismatch between humans and models}

We also found three tasks for which humans do worse when performing verbal thinking, but where this effect does not cleanly translate to models with CoT. One unifying explanation for these effects is humans and models possessing different limitations for task-relevant abilities, such as access to different kinds of information or memory resources.

\paragraph{Explaining a logical inconsistency.} When human participants are shown a pair of logically inconsistent statements and asked to explain their coexistence, they become worse at judging whether the statements are indeed logically inconsistent~\citep{khemlani2012hidden}. In the task, participants are provided with two sentences following the template: ``If $A$ then it is always the case that $B$'', and either ``$A$, but it is not the case that $B$'' or ``It is not the case that $B$''. The former introduces a logical inconsistency, while the latter does not. In one condition humans were first asked to verbally explain why an inconsistent pair could coexist before providing the inconsistency judgement, while in the other they conducted the explanation after the judgement. Performance was significantly worse in the former. 

\begin{table}
\centering
\caption{Comparing zero-shot and CoT on the logical inconsistency task using stimuli from 3 datasets. Random performance is 50\%.}
\label{tab:logic_results}
\resizebox{.48\textwidth}{!}{%
\begin{tabular}{@{}lcccccc@{}}
\toprule
                  & \multicolumn{2}{c}{MNLI} & \multicolumn{2}{c}{SNLI} & \multicolumn{2}{c}{Synthetic} \\
                  & Zero-shot    & CoT       & Zero-shot    & CoT       & Zero-shot       & CoT         \\ \midrule
o1-preview (subset)           & -            & -         & -            & -         & -               & 86.5\%      \\
GPT-4o            & 53.2\%       & 93.9\%    & 51.4\%       & 94.3\%    & 51.0\%          & 74.0\%      \\
Claude 3.5 Sonnet & 65.2\%       & 67.5\%    & 67.4\%       & 69.8\%    & 56.7\%          & 57.8\%      \\
Claude 3 Opus     & 62.7\%       & 58.8\%    & 66.2\%       & 58.7\%    & 54.5\%          & 51.8\%      \\
Gemini 1.5 Pro    & 73.2\%       & 68.2\%    & 68.8\%       & 63.9\%    & 60.5\%          & 61.5\%      \\
Llama 3.1 70B Instruct    & 55.6\%       & 81.6\%    & 50.4\%       & 82.3\%    & 50.0\%          & 65.8\%      \\ \bottomrule
\end{tabular}%
}
\end{table}

The original human experiment contained 12 unique $\{A, B\}$ pairs. To scale this task for LLMs, we leverage existing entailment pairs in natural language inference tasks, which we use as $A$ and $B$ to fill in the sentence templates. We used a combination of three datasets: Stanford Natural Language Inference (SNLI), Multi-Genre Natural Language Inference (MNLI), and a synthetic LLM-generated dataset of 100 entailment pairs. We filtered these datasets for pairs that were labeled ``entailment'' (i.e., $A$ entails $B$). In addition, we limit the format of $A$ and $B$ such that the template forms coherent sentences. In total, we evaluate on 1608 $\{A, B\}$ pairs: 675 from SNLI, 833 from MNLI, and 100 synthetic. Each pair was used to construct two classification problems, one consistent and one inconsistent, for a total of 3216 problems in our dataset. For more details on problem generation see Appendix~\ref{app:logic_generation}. 

We evaluated a suite of state-of-the-art LLMs on this task: o1-preview (on a subset of 30 synthetic questions due to cost constraints), GPT-4o, Claude 3.5 Sonnet, Claude 3 Opus, Gemini 1.5 Pro, and Llama 3.1 70B Instruct. We used one zero-shot prompt and two conditions of CoT: one where the model is simply asked to reason before answering, and another that follows the original experiment by asking the model to explain the inconsistency directly (see Appendix~\ref{app:logic_prompts} for details). Results were very similar across these two conditions, and we report an average over both. 

An important difference between models and humans in this setting is priors over logic: participants in \citet{khemlani2012hidden} were recruited intentionally without logical expertise, but LLMs have access to numerous logical puzzles and knowledge in their training corpuses, creating a synergistic effect with the extra tokens CoT provides. 
As such, CoT often improved performance (see Table~\ref{tab:logic_results}). 
This was especially pronounced in GPT-4o, where CoT improved performance by over 40\% on pairs from MNLI and SNLI. However, in the models that performed best with zero-shot prompting, Gemini 1.5 Pro and Claude 3 Opus, we did see decreases in performance with CoT, potentially mirroring some of the effects in what is seen with humans. 

\paragraph{Spatial intuitions.} Psychologists have documented cases involving spatial reasoning in which humans generate more accurate responses after visual or motor simulation compared to verbal thinking. To investigate whether this applies to models, we replicate a cup-tilting task from \citet{schwartz1999inferences}. In the task, participants are shown an image of two rectangles with varying dimensions, representing two cups --- one empty and one that contains water. Participants are asked to estimate the height of water that should be added to the empty cup so that when tilting both cups, water will reach both rims at the same angle (see Figure~\ref{fig:main_fig}, SI). While the original task had participants draw a water level on the empty cup, LMMs were unable to do this, so we turned the task into a multiple choice question by adding markings $A$ -- $D$ to the side of the empty cup and asking the model to select one. Incorrect options were generated by adding Gaussian noise to the correct answer while satisfying the constraint that options must be a certain distance apart. We scaled up this task by varying the dimensions of cup sizes and water height, creating a total of 100 problems, each with a code-drawn image containing the cups and multiple choice answers (see Appendix~\ref{app:water_generation}). We evaluated with zero-shot and CoT prompts on several open- and closed-source LMMs: GPT-4o, Claude 3.5 Sonnet, Claude 3 Opus, Gemini 1.5 Pro, and InternVL2 Llama3 76B.

In this setting, it is unlikely that large multimodal models would share the same motor simulation capabilities as humans due to lack of representations built from motor experience. The improved performance in the non-verbal thinking condition for humans required spatial or motor intuition, and thus for models we did not observe significant differences between zero-shot and CoT prompts (see Table~\ref{tab:water_results}). Generally, we expect this to extend to other tasks for which models lack task-relevant priors that humans possess.

\begin{table}
\centering
\caption{Comparing zero-shot and CoT on the spatial intuition task.}
\label{tab:water_results}
\resizebox{0.48\textwidth}{!}{%
\begin{tabular}{@{}lccccc@{}}
\toprule
                     & Zero-shot & CoT  & \begin{tabular}[c]{@{}c@{}}Performance\\ (absolute)\end{tabular} & \begin{tabular}[c]{@{}c@{}}Performance\\ (relative)\end{tabular} & $p$-value \\ \midrule
GPT-4o               & 38\%      & 40\% & +2\%                                                                    & +5.00\%                                                             & $0.61 $          \\
Claude 3.5 Sonnet    & 42\%      & 38\% & -4\%                                                                  & -10.53\%                                                              &    $0.28$     \\
Claude 3 Opus        & 42\%      & 38\% & -4\%                                                                    & -10.53\%                                                                &   $0.28$      \\
Gemini 1.5 Pro       & 35\%      & 36\% & +1\%                                                                    & +2.78\%                                                                 &   $0.99$    \\
InternVL2 Llama3 76B & 39\%      & 31\% & -8\%                                                                    & -25.81\%                                                                &    $0.67$    \\ \bottomrule
\end{tabular}%
}
\end{table}

\paragraph{Aggregating features for a decision.} The final category of tasks we consider are tasks that overwhelm human working memory. \citet{dijksterhuis2004think} found that humans made poor choices when deliberating over apartment options if provided with a large amount of information about various apartment features in a short amount of time.
In the study, participants were shown 48 statements for four seconds each, where each statement described a positive, negative, or neutral aspect of one of four apartments. They were then asked to select the best apartment after either deliberating or completing a distractor task. Authors found that the distractor task actually improved performance over deliberating. 

To scale this task up for LLMs, we generated 80 unique apartment features with four statements per feature: one positive, one negative, and two neutral. We then asked GPT-4o to rate the impact each statement would have on the impression of an average tenant in $[-5, 5]$. We randomly sampled apartments by choosing one statement per feature and constructed sets of four where the best apartment had a per-feature average score $\Delta \in \{[0.1, 0.3], [0.3, 0.5], [0.5, 1]\}$ higher than the next-best option. We sampled 300 such sets (100 per $\Delta$ range) to form choice tasks (see Appendix~\ref{app:apartment_generation}). 
We tested several open- and closed-source LLMs with zero-shot and CoT prompts: GPT-4o, Claude 3.5 Sonnet, Claude 3 Opus, and Llama 3.1 70B Instruct. Llama 3.1 70B Instruct was often unable to return an answer after deliberating in the CoT condition, reducing performance. 

In this setting, there were meaningful differences in working memory between humans and models. Humans were forced to rely on their aggregate impressions of each apartment due to the large amount of information. However, even after increasing the number of contextually relevant statements over six-fold, models were able to access all feature statements in-context. Thus, we observed mostly positive effects from CoT (see Table~\ref{tab:apartment_results}). When observing reasoning traces, we found that the availability of context turns the problem into summing up the importances of the features. More capable models were able to leverage CoT to get these approximately correct, despite known issues in long-context retrieval~\citep{liu2023lostmiddlelanguagemodels}, greatly surpassing human memory. This highlights the need to consider fundamental differences in capabilities between models and humans when searching for parallel effects between them.

\begin{table}
\centering
\caption{Results for apartment selection task across four models and three ranges of $\Delta$. }
\label{tab:apartment_results}
\resizebox{0.48\textwidth}{!}{%
\begin{tabular}{@{}lcccccc@{}}
\toprule
\qquad \qquad \qquad $\Delta$             & \multicolumn{2}{c}{{[}0.1, 0.3{]}} & \multicolumn{2}{c}{{[}0.3, 0.5{]}} & \multicolumn{2}{c}{{[}0.5, 1{]}} \\
                  & Zero-shot          & CoT           & Zero-shot          & CoT           & Zero-shot         & CoT          \\ \midrule
GPT-4o            & 47\%               & 45\%          & 57\%               & 56\%          & 80\%              & 87\%         \\
Claude 3.5 Sonnet & 50\%               & 62\%          & 62\%               & 72\%          & 81\%              & 95\%         \\
Claude 3 Opus     & 35\%               & 50\%          & 57\%               & 58\%          & 72\%              & 84\%         \\
Llama 3.1 70B Instruct     &   42\%             &  6\%                 & 44\%               & 5\%           & 43\%              & 20\%         \\ \bottomrule
\end{tabular}%
}
\end{table}

\subsection{Effectiveness of our heuristic}
Our heuristic based on human overthinking identified three tasks with significant CoT-induced performance drops, but mixed or negligible effects were observed in the other three tasks. It is thus reasonable to ask whether selecting tasks based on human behavior reliably isolates tasks where CoT reduces performance, or merely samples randomly from all tasks. To evaluate this, we conducted a bootstrap test using a dataset of 378 comparisons between zero-shot and CoT across different models and tasks from~\citet{sprague2024cot}. From our results, we selected all 50 numerical differences across the six types of tasks across all models\footnote{For the classifying data with exceptions task, we use differences in accuracy reported from Figure~\ref{fig:learning_curve} to match other inputs.}. Resampling 100,000 times with a sample size of 50, none of the 100,000 bootstrapped samples had a greater average decrease in performance than our results, strongly supporting the hypothesis that our heuristic is more efficient than past endeavors at finding failures in CoT based on effect size. 

In addition to effect size, we also test if our heuristic finds a higher occurence of performance decreases (irrespective of magnitude). Using the same bootstrapping procedure and data, we find that only 11 of 100,000 resampled cases yielded equal or greater reductions in performance than our sample (yielding an estimated $p < 0.00011$). These results confirm that our heuristic also disproportionately identifies CoT-induced failures irrespective of effect size. For more detailed explanations of both analyses, see Appendix~\ref{app:effectiveness}.

\section{Discussion}
\label{sec:discussion}

Chain-of-thought prompting is an effective way to expand the capacities of large language and multimodal models. However, knowing that CoT significantly decreases performance in specific settings is important for considering when it should be deployed, and especially whether it should be deployed by default. By using cases where verbal thinking decreases human performance as a heuristic, we successfully identify three settings where CoT results in large decreases in model performance, leading to implications for choosing when CoT should be deployed.

While we draw a connection between human cognition and LLMs/LMMs, we do not claim that these systems operate in the same way or that models should be anthropomorphized. Rather, we see this connection as a tool for identifying settings where the structure of the task or shared limitations result in negative effects of verbal thinking.
Our exploration is guided by considering not only whether verbal thinking reduces human performance, but also whether there are differences between humans and models that must be considered. Our results provide evidence that CoT can result in large decreases in performance when human verbal thinking leads to similar failures, illustrating that we can use the psychology literature to find cases that are informative about the performance of CoT. We now turn to limitations and future directions.

\paragraph{Types of inference-time reasoning.} Since the invention of CoT, researchers have developed new prompting strategies suited for application domains, as well as more elaborate general-purpose prompts such as tree-of-thought~\citep[ToT;][]{yao2023treethoughtsdeliberateproblem} and self-consistency~\citep{wang2023self}. We tested the effectiveness of ToT on GPT-4o for the implicit statistical learning task~(see Appendix~\ref{app:agl_tot}). While ToT improved accuracy (64.55\%~  vs. 62.52\%), this was still far from GPT-4o's zero-shot performance of 94.00\%, suggesting that our findings extend across inference-time reasoning techniques. However, future work is required to determine whether this generalizes to other task domains and methods of eliciting verbal thinking in models. 

\paragraph{Scope of application.} While our psychology-based heuristic offers a strategy for identifying failure cases of CoT, it is unlikely to cover all cases where CoT decreases performance. 
Existing psychological research has been guided by a variety of theoretical and practical considerations, but does not offer an exhaustive or representative sample of all tasks, and will miss cases that are uniquely interesting to study in models but not humans. Thus, we envision our contribution to be complementary to existing evaluation methods in natural language processing.

As we've seen across our six tasks, knowledge of what drives a decrease in performance in humans can be leveraged to generate predictions about the effects of CoT, but this remains an inferential step that requires careful reasoning and an understanding of model capabilities. Despite these limitations, our method can be used to identify large and consequential failures of CoT, as documented in our three failure cases. It also offers valuable cross-domain insight that can help build intuitions and contribute to our overall understanding of inference-time reasoning. 
On the flipside, the existence of capable LLM/LMM systems also allows us to better understand why human performance can be degraded by deliberation. By considering when CoT's effects mirror humans and when they do not, we can distinguish when the task or mechanisms shared by humans and models are responsible for failures, versus when the issues arise from uniquely human strategies or limitations. 
\paragraph{Alternative explanation for mismatch between CoT and humans. }
Another explanation for why we do not see consistent drops in performance in the latter three tasks is that how we implemented the tasks for LLMs removed the failure effect. It’s possible that with other implementations (e.g., using a memory-based LLM agent for the apartments task) we might in fact see decreased performance mirroring humans. While we explored prompt variations for each task, these were not exhaustive due to the endless space of changes to prompts. In other words, because the tasks were inevitably changed to scale up the evaluation and match more realistic applications of models, it’s also possible that these changes are what explain the human-model mismatch. 

\paragraph{Future directions.}
While we have focused on CoT reasoning, the framework presented here suggests a more general strategy for leveraging empirical work on humans to characterize the performance of models: jointly considering (i) psychological results concerning humans and (ii) whether the relevant constraints that shape human performance extend to those models. For example, it could be fruitful to investigate effects of comparison or analogical prompting \citep{lombrozo2024learning} through this lens. 

Overall, we envision studying how to evaluate and improve models as a collaborative effort between machine learning methods, psychological insights, and a burgeoning literature connecting and comparing humans and models. By sharing knowledge and building strong collaborations between these disciplines, we can utilize rich insights from decades of studying humans to advance the domain's intuitions about models and analyze an even broader array of tasks and applications for AI. 

\newpage
\section*{Acknowledgments}
Experiments were supported by Azure credits from a Microsoft AFMR grant. This work and related results were made possible with the support of the NOMIS Foundation. Special thanks to Howard Chen, Jian-Qiao Zhu, and Mengzhou Xia for providing invaluable feedback on this project. 

\section*{Impact Statement}

This paper presents work whose goal is to advance the understanding of inference-time reasoning between humans and large language and multimodal models. While there are many potential societal consequences 
of our work, we are explicit about the takeaways we intend for our paper (see Section~\ref{sec:discussion}) and thus we don't forsee that any additional details need to be highlighted here. 

\bibliography{iclr2025_conference}
\bibliographystyle{icml2025}

\newpage
\appendix
\onecolumn
\section{Implicit Statistical Learning Task}
To study cases involving implicit statistical learning, we consider an artificial grammar learning task. In the task, LLMs are provided with letter string train examples that belong to the same category, and are tasked to classify whether a new string belongs to the same category. 

\subsection{Generation of Artificial Grammar Learning Dataset}
\label{app:agl_generation}
\begin{figure}[b!]
    \centering
    \includegraphics[width=0.6\linewidth]{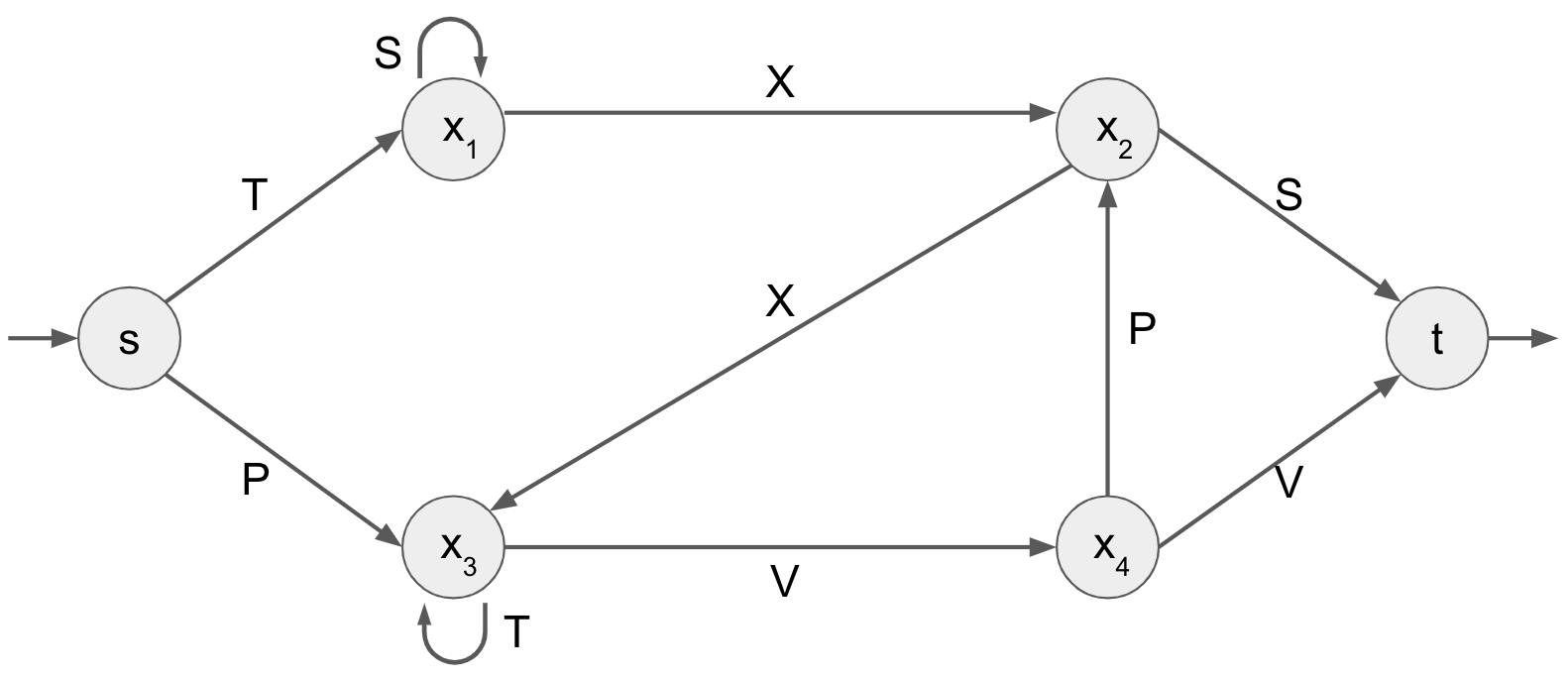}
    \caption{The FSG used in \citet{fallshore1993post} and \citet{reber1977implicit}, two classic studies on artificial grammar learning. This FSG was used the generate strings for all participants in both studies. We form our dataset using FSGs that follow a similar structure.}
    \label{fig:fsg_original}
\end{figure}

In the original psychology experiments~\citep{fallshore1993post, reber1977implicit}, participants performed the classification task on strings generated by a fixed finite state grammar (FSG) constructed by the researchers (see Figure~\ref{fig:fsg_original}). A string is generated by the FSG if it corresponds to a valid path along the directed edges from the source node $s$ to the sink node $t$, where the letters on the path are appended together. 

\begin{figure}
    \centering
    \includegraphics[width=0.61\linewidth]{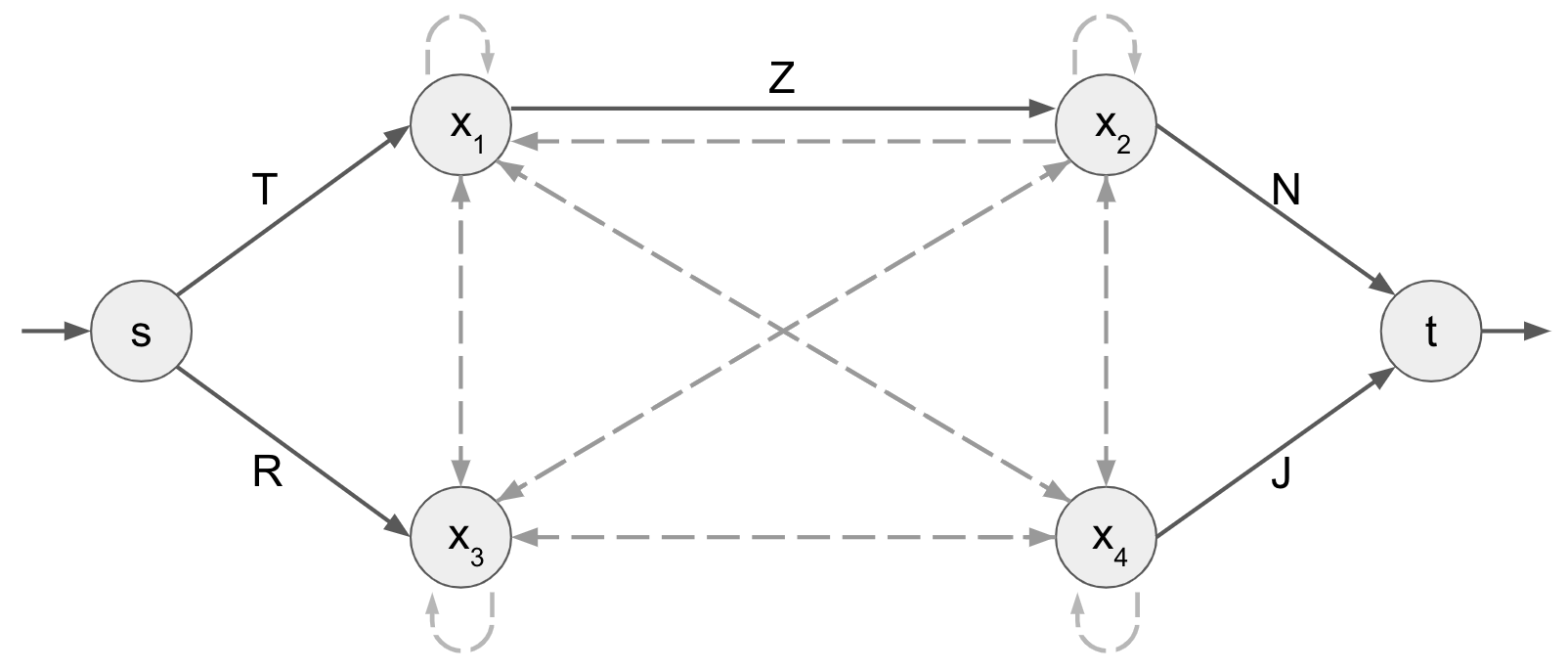}
    \caption{The potential FSGs used in our dataset. Directed edges that always exist are in black, while the others that could exist are dashed and in gray. Bi-directional arrows denote two potential directed arrows. The letters on each edge represent a random sample.}
    \label{fig:fsg_ours}
\end{figure}

In our experiments, we expand the experiments massively to 100 randomly sampled FSGs that follow the same rough structure of those used in the experiment. 
To scale up the dataset, we construct and sample from all possible FSGs that obey the following rules. For a visual representation please see Figure~\ref{fig:fsg_ours}.
\begin{itemize}
    \item 6 nodes total, including source $s$, sink $t$, and four nodes $x_1, \dots, x_4$.
    \item Edges $(s, x_1), (s, x_3), (x_2, t)$ and $(x_4, t)$ are always present.
    \item Edge $(x_1, x_2)$ is always present to avoid isomorphisms and the null case where no paths exist from $s$ to $t$.
    \item The remaining middle edges $\{(x_1, x_3)$, $(x_1, x_4)$, $(x_2, x_1)$, $(x_2, x_3)$, $(x_2, x_4)$, $(x_3, x_1)$, $(x_3, x_2)$, $(x_3, x_4)$, $(x_4, x_1)$, $(x_4, x_2)$, $(x_4, x_3)\}$ can either exist or not, for a total of $2^{11}$ combinations.
    \item Each $x_i$ can have self-loops, e.g., $(x_1, x_1)$, for a total of $2^4$ combinations.
    \item Letters on each edge are randomly selected from the capital alphabet, for a total of $26^8$ combinations.
    \item Each FSG should be able to generate at least $37$ unique strings with length $\leq 8$.
    \item The construction of the FSG is unique with respect to the three graphical isomorphisms that each FSG satisfying the rules could have. 
\end{itemize}

For each FSG, we sampled paths of up to length 8 and used them as stimuli for the experiment. Following \citet{fallshore1993post}, we sampled $37$ to use in the experiment, assigning 15 to be training examples and $22$ to be positive test examples. We also constructed $22$ negative test examples by sampling a random string from the FSG, perturbing one letter in a randomly selected position to another letter that exists on some edge of the FSG. We ensured that the negative examples did not belong to the FSG. 

In total, this yielded 4400 individual questions asked to the large language models. Each question was asked individually after the 15 training examples. See the next section for the specific prompts. 

\subsection{Prompts}
\label{app:agl_prompts}

For our experiments, we prompted the models using one zero-shot prompt and one CoT prompt. The zero shot prompt is shown in Table~\ref{tab:agl_prompt}. For Claude, GPT, and Gemini models, we use temperature = $0.0$. For o1, the beta version limited its usage to temperature = $1.0$. For open-source models, we use temperature $= 0.0$. Max tokens was set to 10 for zero-shot and 1000 for CoT. The remaining hyperparameters were set at their default values: top\_p, top\_k, seed, min\_tokens, etc.

\begin{table}[ht]
\centering
\caption{Example prompt for artificial grammar learning task, zero shot. }
\label{tab:agl_prompt}
\resizebox{0.79\textwidth}{!}{%
\begin{tabular}{l}
\hline
\textbf{Prompt:} \\
\begin{tabular}[c]{@{}l@{}}These strings were generated according to a certain set of rules. Does the following string also \\follow the same set of rules?\\ 
$[$test example$]$ \\ 
Please ONLY answer ``Yes" or ``No".
\end{tabular} \\ \hline
\end{tabular}
}
\end{table}

The CoT prompt uses one of the most common prompting methods for chain-of-thought, replacing the last line with, `Please reason about your answer before answering ``Yes" or ``No".'

When conducting pilot experiments, we also tried a version of the prompts where we asked models to ``memorize the following letter strings'' in the first line of the prompt as this was more in-line with the original human experiment. We found that results were extremely similar to the more general version shown above, and thus discarded this more specialized case. 

\subsection{CoT Failure Example}
An example CoT prompt and output where GPT-4o fails for the artificial grammar learning task is in Table~\ref{tab:agl_cot_example}.

\begin{table}[ht]
\centering
\caption{Example CoT prompt and GPT-4o output for artificial grammar learning task.}
\label{tab:agl_cot_example}
\resizebox{\textwidth}{!}{%
\fontsize{12}{14}\selectfont %
\begin{tabular}{l}
\hline
\begin{tabular}[c]{@{}l@{}}
\textbf{Prompt:} \\
Consider the following letter strings: \\
FDMQUQC FDQMUMR DZGBDMR DGUMQQR DGUMUMR
FDXDC FDMUXDC DGUMQR FDQXDC DGBDC\\
FDMQQR DGQQR DZZGR DGQQUC FDMUMR\\ \\
These strings were generated according to a certain set of rules. Does the following string also follow the same set of rules?\\
DGBDQC\\
Please reason about your answer before answering ``Yes" or ``No". \\ \\
\textbf{Model output: }\\
To determine whether the string ``DGBDQC" follows the same set of rules as the given strings, we need to analyze the \\patterns and rules that might be present in the provided strings.\\ \\
Let's break down the given strings to identify any patterns:\\ \\
1. **Length Consistency**: All strings have a length of 7 characters.\\
2. **Character Positions**:\\ 
    - The first character can be either `F' or `D'.\\
    - The second character can be `D', `G', or `Z'.\\
    - The third character can be `M', `Q', `G', `X', or `Z'.\\
    - The fourth character can be `Q', `U', `X', `B', or `D'.\\
    - The fifth character can be `Q', `U', `X', `B', or `D'.\\   - The sixth character can be `C', `R', or `D'.\\   - The seventh character can be `C', `R', or `D'. \\ \\
    3. **Repetition and Pairing**:\\   - Some strings have repeated characters, such as ``FDMQUQC" and ``DGUMQQR".\\   - Some strings have specific pairs of characters that appear together, such as ``UM" in ``DGUMUMR" and ``FDMUMR".\\ \\
    4. **Character Sets**:\\   - The characters used are from the set {D, F, G, M, Q, R, U, X, Z, C, B}.\\ \\
    Now, let's check the string ``DGBDQC":\\ \\1. **Length**: The string ``DGBDQC" has 6 characters, which is different from the 7-character length of the given strings.\\2. **Character Positions**:\\   - The first character `D' is valid.\\   - The second character `G' is valid.\\   - The third character `B' is valid.\\   - The fourth character `D' is valid.\\  - The fifth character `Q' is valid.\\   - The sixth character `C' is valid.\\   - However, the string is missing a seventh character.\\ \\
    Given that the string ``DGBDQC" does not match the 7-character length rule observed in all the provided strings, it does not \\follow the same set of rules. \\ \\
    
    Therefore, the answer is **\textbf{No}**. \\ \\ 
\textbf{Ground truth: }\\ 
Yes
\end{tabular} \\ \hline
\end{tabular}
}
\end{table}

\subsection{Tree-of-thought Experiments}
\label{app:agl_tot}

\begin{table}[ht!]
\centering
\caption{Results comparing zero-shot, CoT, and ToT on a subset of the artificial grammar learning task.}
\label{tab:agl_tot}
\resizebox{0.8\textwidth}{!}{
\begin{tabular}{lccccccl}
\hline
                  & Zero-shot & CoT & ToT    & \begin{tabular}[c]{@{}c@{}}Performance\\ decrease (CoT)\end{tabular} & \begin{tabular}[c]{@{}c@{}}Performance\\ decrease (ToT)\end{tabular} & \begin{tabular}[c]{@{}c@{}}$p$-value\\(CoT)\end{tabular} & \begin{tabular}[c]{@{}c@{}}$p$-value\\(ToT)\end{tabular}                \\ \hline
\multirow{2}{*}{GPT-4o}   & \multirow{2}{*}{94.00\%}  &\multirow{2}{*}{62.52\%} & \multirow{2}{*}{64.55\%} &\multirow{2}{*}{$31.48\%$}   & \multirow{2}{*}{$29.45\%$} & \multirow{2}{*}{$<0.0001$} & \multirow{2}{*}{$<0.0001$}\\

               \\ \hline
\end{tabular}
}
\end{table}

To analyze whether our hypotheses about model chain-of-thought extend to other types of inference-time reasoning, we evaluated the performance of GPT-4o with tree-of-thought (ToT)~\citep{yao2023treethoughtsdeliberateproblem} on a subset of 10 artificial grammars, totaling 440 examples. Given the input prompt, we asked the LLM to generate five different thoughts to explain whether the string also followed the same set of rules as the in-context examples. Then, five votes were conducted to select the best thought, which we parsed and compared to the ground truth label. We found that ToT resulted in a small improvement over the task (64.55\% vs. 62.52\% accuracy, see Table~\ref{tab:agl_tot}), but this performance was still much worse compared to GPT-4o zero-shot accuracy (94.00\%). This suggests that the reduction in performance is not only associated with CoT, but also other types of inference-time reasoning. 

\subsection{FSG Complexity Reduction Experiments}
\label{app:agl_complex}

\begin{figure}
    \centering
    \includegraphics[width=0.5\linewidth]{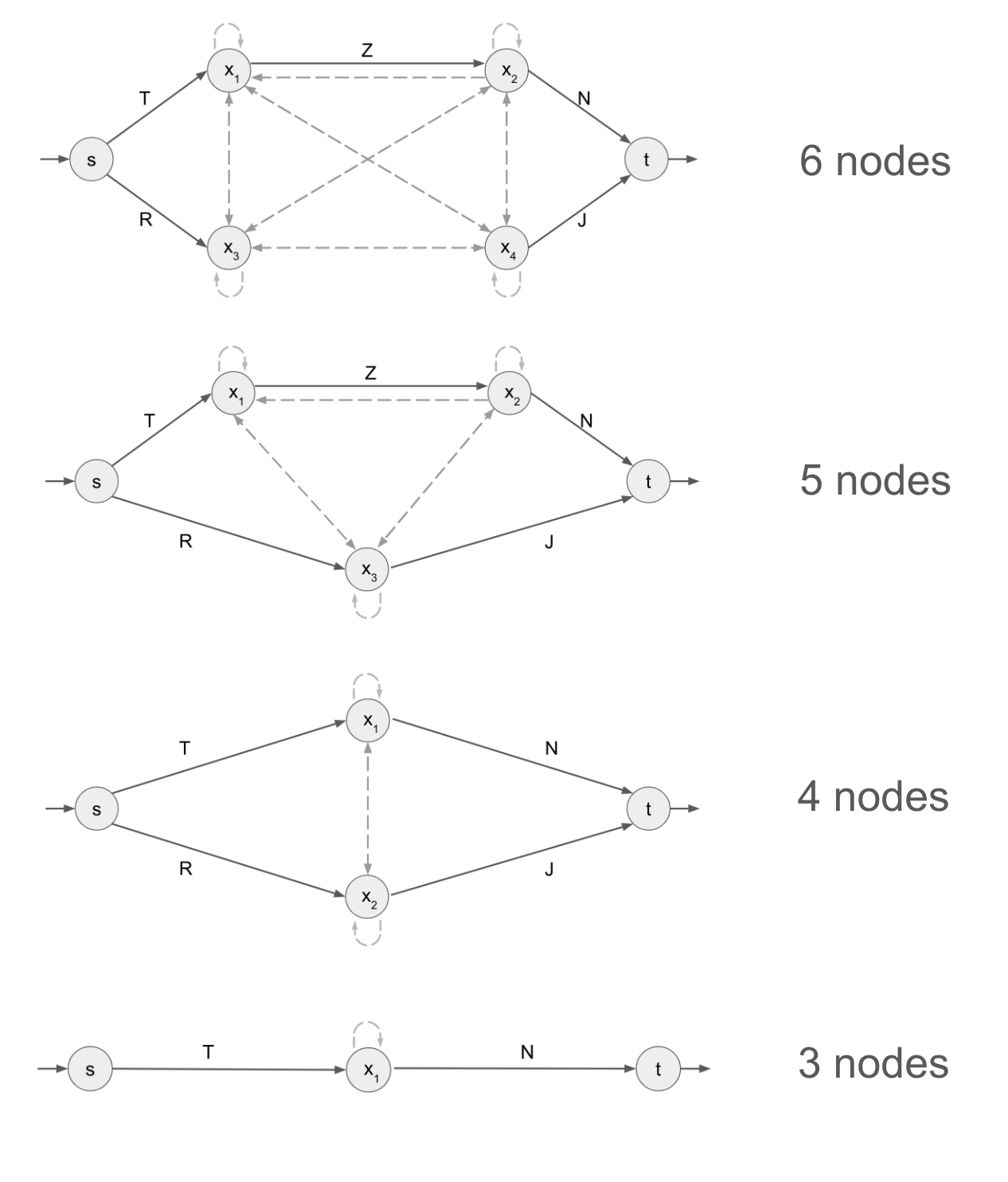}
    \caption{Different complexities in the underlying finite state grammar for implicit statistical learning.}
    \label{fig:agl_complex}
\end{figure}

To show generalizability across variations of each task, we conduct additional experiments varying the problems in each of the failure cases. 
For artificial grammar learning, we varied the complexity of underlying FSG. Specifically, we reduced the nodes in the finite state automata that generated the artificial grammars. While the original had 6 nodes, we iteratively reduced to 5, 4, and 3 nodes using edge contraction (see Figure~\ref{fig:agl_complex}). Across all valid FSGs with no unused nodes, we observed the following accuracies:

5 nodes, zero-shot = 0.886, CoT = 0.766

4 nodes, zero-shot = 0.837, CoT = 0.665

3 nodes, zero-shot = 1.000, CoT = 1.000

Excluding the reasonably trivial case of three nodes, where we could not detect a difference between the zero-shot and CoT conditions, our results are consistent across variations in difficulty. This provides further confidence in the generalizability of the findings, that CoT hurts performance not in just a one-off task, but in a generalizable class of tasks in implicit statistical learning. 

\subsection{Robustness to Sampling Temperature}
\label{app:agl_temp}
To further test the robustness of our results, we conducted ablatons on different temperatures (t = 0.5, 0.7) across the full 4400 problems for the artificial grammar learning task on GPT-4o. The resulting accuracies were as follows:

t = 0, zero-shot = 87.5, CoT = 64.4 (original results for reference)

t = 0.5, zero-shot = 88.3, CoT = 63.6

t = 0.7, zero-shot = 87.8, CoT = 63.6

Thus, our results seem to be robust to variations in temperature sampling.

\subsection{In-context Steering  to Improve CoT for Artificial Grammar Learning}
\label{app:agl_steering}
To study whether explicitly prompting the model to focus on an optimal reasoning method improves performance, we conducted additional experiments using the same subset of problems as evaluated with the OpenAI o1, assessing performance on GPT-4o.
Specifically, the prompts were designed to explicitly instruct the model to identify and utilize bigram patterns when classifying words as belonging to or not belonging to FSGs, which is advantageous because of the nature in which accepting strings are generated. Under these conditions, we obtained the accuracy of 68.43\%, compared to 87.50\% in the zero-shot condition and 64.40\% in the CoT condition. Consistent with our expectations, this result indicates that while explicit bigram instruction did improve CoT performance, it still did not reach that of zero-shot prompting. 
\section{Facial Recognition Task}
To study tasks where language impairs the recognition of visual stimuli, we focus on a facial recognition task, where VLMs are asked to select one of five candidate images that matches the face of a provided image. The original experiment in \citet{schooler1990verbal} had participants view a 30-second video of an individual robbing a bank and then perform a 20-minute distractor task, before either writing down descriptions of the robber's face or doing a distractor task for 5 minutes. Participants were then provided with 8 verbally similar faces to choose from, and those who performed the written description performed much worse (38\% vs. 64\% accuracy) at identifying the robber. 

\subsection{Generation of Facial Recognition Dataset}
\label{app:face_generation}

To adapt this task to testing models, we made a few design decisions. First, we chose to replace the initial video stimuli with an image of the person's face to allow for the testing of vision language models. Next, we chose to remove the distractor tasks. This decision was based on pilot results indicating that common psychology distractor tasks such as the n-back task~\citep{kirchner1958age} resulted in large amounts of noise in model outputs, while other distractors were of limited effect on the model due to it being able to retrieve the earlier stimuli in-context. Furthermore, even without the distractor, models already showed a large difference in performance across zero-shot and CoT conditions. 
Thus, our task was simplified to a facial matching task, where a model was given a human face as input and responded with the index of the matching face image as its output. 

To generate the faces for the facial recognition dataset, we use stable-image-ultra~\citep{stabilityai_api_2024}. We experimented with other models such as DALL-E 2~\citep{openai_dalle2} and DALL-E 3, but found generation capabilities were significantly less realistic than stable-image-ultra. This difference was especially pronounced in generating realistic facial images of people in racial minorities. 

To cover a diverse set of human faces, we prompt models to generate faces with features age \{young, middle-aged, old\}, race/ethnicity \{asian, black, hispanic, white\}, gender \{man, woman\}, eye color \{brown, blue, green\}, hair color \{brown, black, blonde, red, gray\}, hair length \{long, short\}, and hair type \{curly, wavy, straight\}. We removed some low-probability combinations such as red hair with asian ethnicity due to poorer quality of image generation. Then, we randomly sampled combinations of features to form a descriptor set. 

One issue with stable-image-ultra is that when asked naively to generate an image of the same person as another image, it would alter some details such as ear shape, nose shape, or other facial ratios that would make it impossible to be the exact same person. We addressed this issue by prompting the stable-image-ultra image generation model to 
\begin{quote}
    ``Generate two realistic images of the same person, one on the left and one on the right. The person should have the following description: [description]''.
\end{quote}
After doing so, we were able to manually check and verify that the faces shown in the two images is clearly the same to the naked eye. One of these images was assigned to be the initial stimuli shown, while the other would be shuffled into the list of answers. 

We also ensured that the other remaining images were 1) clearly not of the same person as the image, and 2) the pose of the person, which was often similar between the pair of generated images, was also replicated in the other fake answers. 
This was achieved using the following prompt with the \textit{edit structure} task in the set of image control API calls from StabilityAI:
\begin{quote}
    ``Generate an image of a \textit{unique} person with the same pose and style as the image provided. The person should have the following description: {description}''.
    
    image input: [correct answer image]
\end{quote}

\begin{figure}
    \centering\includegraphics[width=0.75\linewidth]{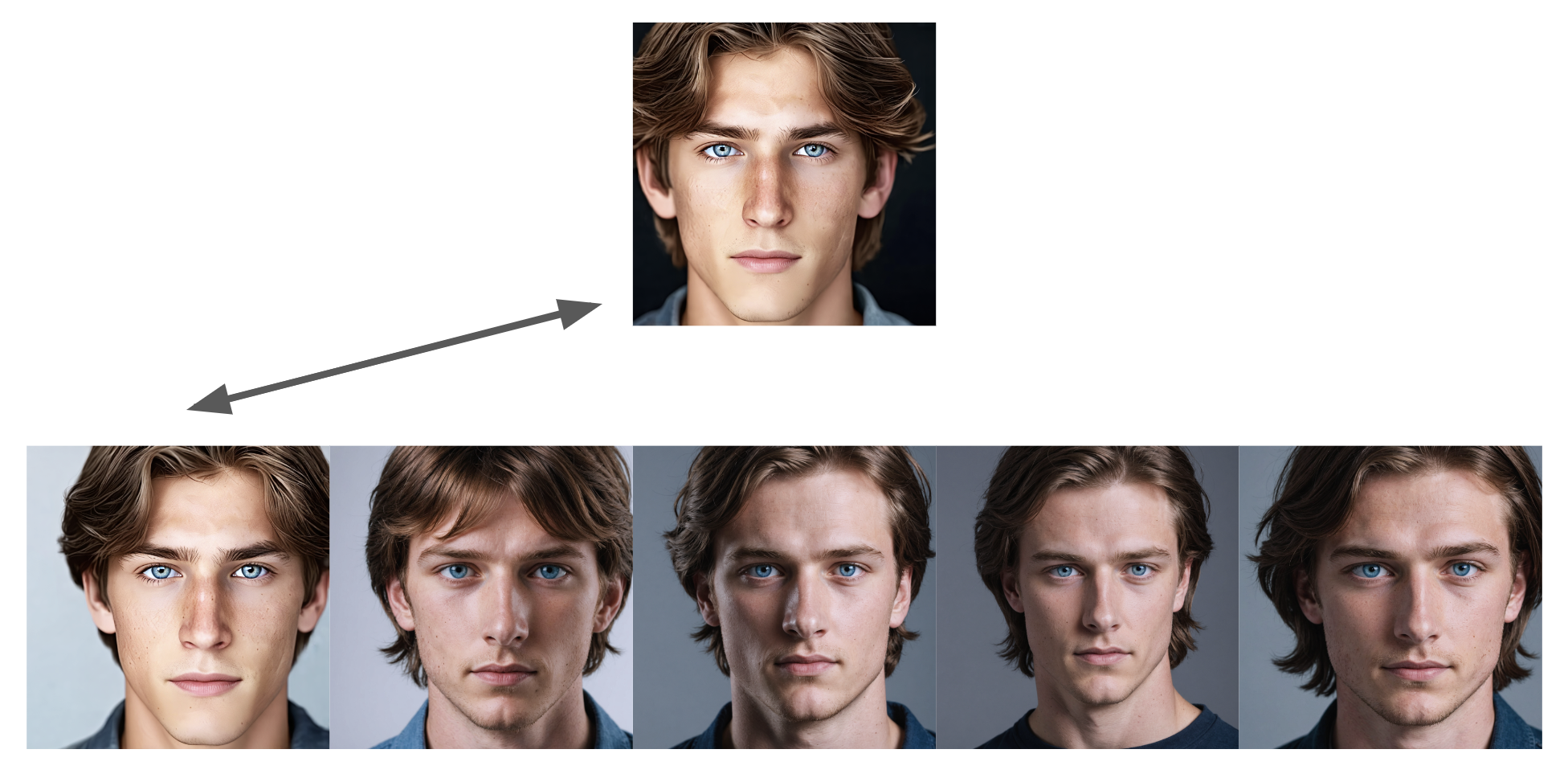}
    \caption{An example of the six images generated for a problem. The first row contains one of the pair of generated images. The first image in the second row contains the other image in the pair, and the remaining four images are incorrect answers generated from this image. }
    \label{fig:face_example}
\end{figure}

Once all the answers were generated, we manually verified the quality of generated images, and ensured that each of 1) and 2) were satisfied. An example of the images generated for a problem are shown in Figure~\ref{fig:face_example}. 

\subsection{Prompts}
\label{app:face_prompts}
To evaluate models on the facial recognition task, we used one zero-shot prompt and one CoT prompt. The zero shot prompt is shown in Table~\ref{tab:vision_prompt}. For all models, we use temperature = $0.0$. Max tokens was set to 10 for zero-shot and 1000 for CoT. The remaining hyperparameters were set at their default values: top\_p, top\_k, seed, min\_tokens, etc.
\begin{table}[ht]
\centering
\caption{Example prompt for facial recognition task, zero shot. }
\label{tab:vision_prompt}
\resizebox{0.64\textwidth}{!}{%
\begin{tabular}{l}
\hline
\textbf{Prompt:} \\
\begin{tabular}[c]{@{}l@{}}Here is an image of a person. \\
$[$image of initial person$]$\\ 
Select the image that contains the same person as the person in the first image.\\
$[$five images of possible matching faces$]$ 
\end{tabular} \\ \hline
\end{tabular}
}
\end{table}

The CoT prompt uses the most original chain-of-thought prompting method by appending ``Let's think step by step'' to the end of the zero-shot prompt, with no other changes.

\subsection{CoT Failure Example}
An example CoT prompt and output where GPT-4o fails for the facial recognition task is in Table~\ref{tab:face_generation_example}. 

\subsection{Facial Recognition Difficulty Reduction Experiment}
\label{app:face_difficult}

\begin{figure}
    \centering
    \includegraphics[width=0.75\linewidth]{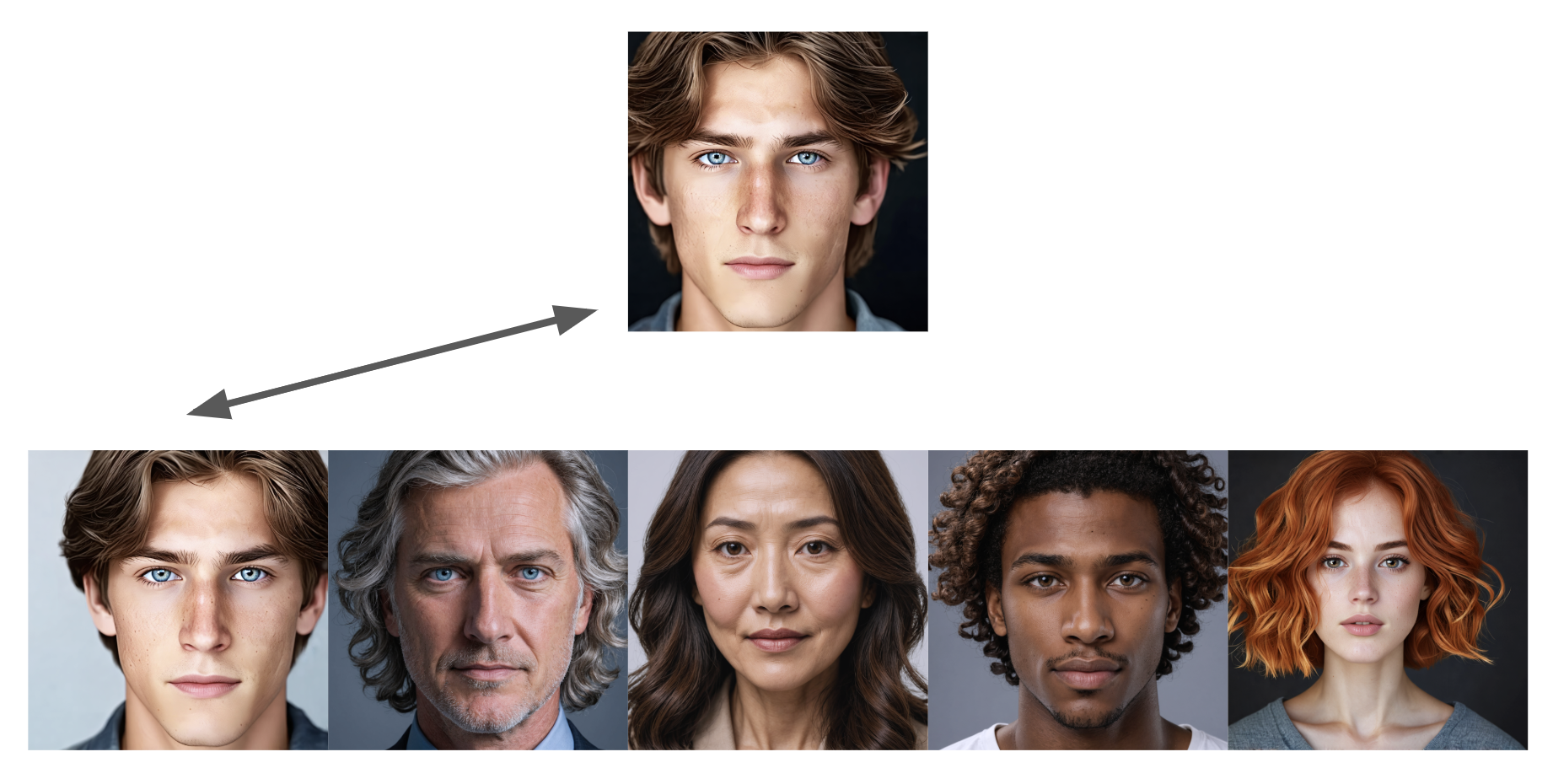}
    \caption{Reduced difficulty in the facial recognition task; images are from the original corpus but do not match the same description.}
    \label{fig:face_difficult}
\end{figure}

To show generalizability across variations of each task, we conduct additional experiments varying the problems in each of the failure cases. 
For the facial recognition task, we relax the constraint that the faces need to have the same verbal description, and instead  we sample 5 faces with different descriptions to form each recognition problem. For a visual representation, see Figure~\ref{fig:face_difficult}.

Across 100 randomly selected sets within this easier paradigm, we find that CoT continues to drastically reduce model performance. GPT-4o has a direct prompting accuracy of 0.61, but CoT accuracy is only 0.32, corroborating our findings that CoT reduces performance across facial recognition tasks.

\subsection{Does CoT distort surface-level prerequisites instead of reasoning?}
\label{app:face_order}

One possible explanation for CoT reducing LMM performance is that verbal processing distorts the ordering of the image options. In order to determine whether CoT prompting introduces confusion related to image ordering, we conducted a simpler version of the facial recognition task with just binary classification. We evaluate GPT-4o on 100 facial recognition problems, each consisting of only two images, and prompt the model to judge whether they represented the same person. The dataset we test on consists of exactly half positive ("yes") and half negative ("no") answers.

In this simplified setting, GPT-4o achieved an accuracy of 72\% under zero-shot prompting. However, accuracy decreased to 62\% when using CoT prompting. These results align with our hypothesis that CoT can impair performance, even in a simplified binary classification version of the task, suggesting that CoT disrupts the reasoning process rather than surface-level problem attributes in facial recognition problems.

\begin{table}[ht]
\centering
\caption{Example CoT prompt and GPT-4o output for facial recognition task.}
\label{tab:face_generation_example}
\resizebox{0.98\textwidth}{!}{%
\fontsize{12}{14}\selectfont %
\begin{tabular}{l}
\hline
\begin{tabular}[c]{@{}l@{}}
\textbf{Prompt:} \\
Here is an image of a person. \\ \\
\includegraphics[width=0.15\textwidth]{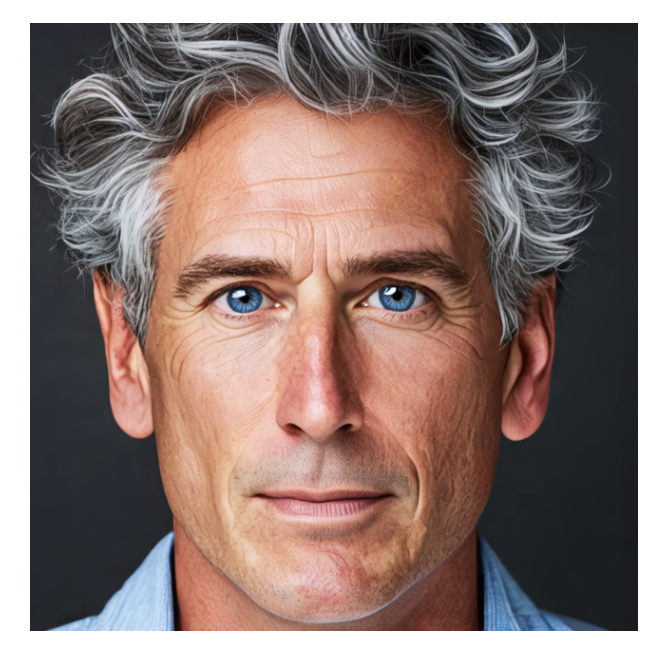} \\
Let's think step by step before selecting the image that contains the same person as the person in the first image. \\ 
\includegraphics[width=0.8\textwidth]{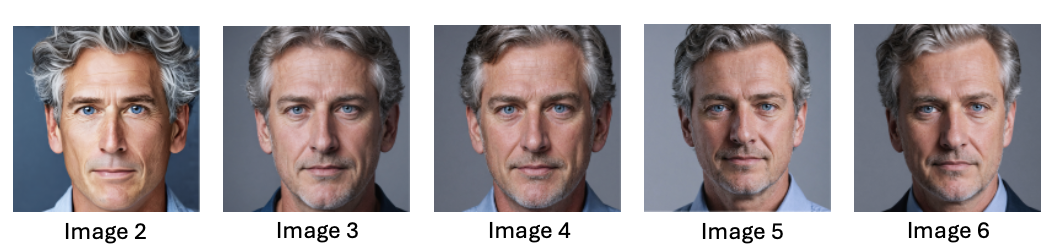}
\\ 
\textbf{Model output: }\\
    To identify the person from the first image, closely compare distinct features such as hairstyle and the visible part of clothing. \\ 
    The first and the third image show the same ear shape and hairstyle. Hence, the same person appears in the \textbf{third image}. \\ \\

\textbf{Ground truth: }\\
Image 2 (2nd row, leftmost)\\

\end{tabular} \\ \hline
\end{tabular}
}
\end{table}
\section{Data with exceptions task}
In this task, we analyze the effect that CoT prompting has on the ability of LLMs to learn a classification of objects that appear to follow a pattern, but with exceptions. In these types of settings, \citet{williams2013hazards} reveal that when humans are given opportunities to deliberate after receiving feedback, they learn more slowly and make more errors compared to those who do not deliberate. The active form of thought mentally ingrains incorrect patterns that shift when exposed to successive unexpected answers, altogether leading to the creation of many deceptively incoherent lines of reasoning throughout the learning process that hinder the ability to directly keep track of the correct labels even after multiple passes.

\subsection{Vehicle dataset generation}
\label{app:vehicle_generation}

We build off of the experimental set-up in \citet{williams2013hazards} where in each trial, we first create a list of objects (vehicles) that are either warm- or cold-climate, which is the label which we want models to learn. Based on this label, we generate one feature that correlates with this target label completely (see Column 2 of Table~\ref{tab:vehicle_features}), and flip this 20\% of the time to create exceptions in the data. 

In addition to this discriminating feature, following~\citet{williams2013hazards}, we also include 1) one unique feature which is different for each object and 2) three additional features whose values are randomized and have no connection with the object class. The unique feature in the original experiment was vehicle color, which we replaced with the license plate for realism. An example setup is depicted in Table \ref{tab:vehicle_features}.

\begin{table}[ht]
\centering
\resizebox{0.8\textwidth}{!}{%
\begin{tabular}{@{}lllllll@{}}
\toprule
\multicolumn{2}{c}{\textbf{Unique features}} & \multicolumn{2}{c}{\textbf{Pattern-related features}} & \multicolumn{3}{c}{\textbf{Irrelevant features}} \\
\cmidrule(r){1-2} \cmidrule(r){3-4} \cmidrule(r){5-7}
& \textbf{License Plate} & \textbf{`Cold' (Class A)/`Warm' (Class B) climate} & & \textbf{Transmission} & \textbf{Seat covers} & \textbf{Doors} \\
\midrule
& A23BCD    & Drives on glaciers        & & Manual   & Cloth & Two \\
& B34EFG    & Made in Norway            & & Automatic & Vinyl & Two \\
& C45HIJ    & Used in mountain climbing & & Automatic & Vinyl & Four \\
& D56KLM    & \textbf{Drives in jungles}        & & Manual & Vinyl & Four \\
& E67NOP    & Has treads                & & Manual & Cloth & Two \\
\midrule
& F78QRS    & \textbf{Heavily insulated}         & & Manual   & Vinyl & Four \\
& G89TUV    & Made in Africa            & & Manual & Cloth & Four \\
& H90WXY    & Has wheels                & & Automatic & Cloth & Two \\
& J12ZAB    & Lightly insulated         & & Manual & Vinyl & Two \\
& K23CDE    & Used on safaris           & & Automatic & Vinyl & Two \\
\bottomrule
\end{tabular}
}
\caption{Sample vehicle classification list. Boldened features indicate flipped labels that break the initial classification pattern.}
\label{tab:vehicle_features}
\end{table}

We sampled 240 sets of 10 vehicles each and prompt the model to learn the labels of the vehicles in a multi-turn setting, which we detail below. 

\subsection{Prompts}
\label{app:vehicle_prompts}
Models are provided with text descriptions of a vehicle's features one vehicle at a time, iterating through the full set of ten vehicles repeatedly up to 15 times. Each time the model is given a set of features, it predicts the corresponding label and subsequently receives feedback for its answer. In contrast to previous experiments, the problems, the model's previous guesses, and the feedback given to the model are all stored in-context and provided to the model in its next prediction. 

In each iteration, the vehicles' order shown to the participant is shuffled. Prompting stopped when the model correctly classified all of the vehicles in one iteration, or reached 15 iterations without performing this successfully. We used one zero-shot prompt and one CoT prompt. The zero-shot prompt was as follows in Table~\ref{tab:vehicle_prompt}. 

\begin{table}[ht]
\centering
\caption{Example prompt for vehicle classification task, zero shot. }
\label{tab:vehicle_prompt}
\resizebox{0.7\textwidth}{!}{%
\begin{tabular}{l}
\hline
\textbf{[Chat history including previous prompts, model predictions, and feedback]}\\
\\
\textbf{Prompt:} \\
\begin{tabular}[c]{@{}l@{}}
The vehicle description is as follows:\\
License plate: [license plate]\\
Descriptor: [descriptor]\\
Transmission: [transmission]\\
Seat Cover: [seat cover]\\
Doors: [doors]\\
Is this vehicle more likely to be a Class A or Class B vehicle? Only answer with `A' or `B'.
\end{tabular} \\ \hline
\end{tabular}
}
\end{table}

In the CoT condition, instead of replicating the human study and asking the model to deliberate after each piece of feedback, we modify the prompt asking the model to make a prediction. Specifically, we replace ``Only answer with `A' or `B'." with 
``Let's think step by step and answer with either `A' or `B'. If you are unsure, feel free to guess and explain your reasoning''.

We append the last sentence because we observed that sometimes the model would refuse to answer based on lack of information.
While we could have also implemented deliberations after each feedback to stay more faithful to the human experiment, our ultimate goal is to inform chain-of-thought, and CoT is most often applied during the process of asking questions to the model rather than having it reflect by itself. 
Furthermore, we believe that these settings are approximately equivalent: Deliberation in human experiments would focus on explaining the feedback provided, but this is also the case in this paradigm because the model would perform reasoning on the previous feedback provided when performing CoT during the prediction of the next label. 

For all models, we use temperature = $0.0$. Max tokens was set to 10 for zero-shot and 1000 for CoT. The remaining hyperparameters were set at their default values: top\_p, top\_k, seed, min\_tokens, etc.

\subsection{Per-round accuracy analysis}
\label{app:vehicle_analysis}

Figure \ref{fig:learning_curve} depicts the aggregate accuracy (correctly predicted examples out of 10) of GPT-4o with direct and CoT prompts over 15 iterations through the list. Although CoT performs better than direct on the first iteration of the list, direct prompting quickly surpasses the performance of CoT by attaining perfect classification ability on the third iteration. Chain-of-thought prompting stagnates in performance at an accuracy level equivalent to the percentage of exemplars whose class designation adheres to the corresponding first-glance generalizable rule (80\%). This suggests that the verbal thinking of CoT biases the model towards predicting via generalizable rules, even when there are more useful features that map exactly to correct answers in context. 

It is worth noting that CoT's tendency towards generalizable rules is often very helpful in other settings. For example, CoT does benefit from this tendency in the predictions of the first pass when all stimuli are previously unseen. This is in line with our conclusion that different strategies for prompting should be chosen based on the task, and neither is always better than the other. 


\subsection{CoT Failure Example}
An example CoT prompt and output where GPT-4o fails for the classifying data with exceptions task is in Table~\ref{tab:vehicle_cot_example}.

\subsection{Robustness Analysis: Reducing Difficulty via Binary Oracle}
\label{app:vehicle_difficulty}

We analyze how we can steer GPT-4o to attain consistently perfect list classification in fewer rounds with chain-of-thought but at the same time, steering it to focus less on the distracting factor, namely, the pattern-related features as depicted in Table \ref{tab:vehicle_features}.

Initially, we performed an environmental intervention where we injected an oracle feature for each vehicle that directly mapped to the vehicle type (Class A or Class B) by altering the license plate feature to be `license plate type', a binary feature (either `0' or `1') instead of the original randomized six-character-long alphanumeric string. While an inconsistency between the pattern-related feature (hot climate or cold climate vehicle) and the vehicle type was still present as formed initially, Class A vehicles always had a license plate type of `0', and Class B vehicles always had a license plate type of `1'. Direct prompting was able to consistently achieve perfect list classification in an average of 1.84 iterations -- in several cases, direct prompting was able to achieve perfect classification on the first iteration due to an initial correct guess by chance that likely allowed the LLM to internalize the direct connection between license plate type and vehicle class. However, CoT prompting still performed considerably worse, failing to achieve perfect classification in the maximum number of iterations alloted (7) in 29 out of 30 trials. Notably, CoT prompting only allowed the LLM to pick up on the relationship between license plate type and vehicle class in only one trial, allowing the LLM to achieve perfect list classification on the second iteration.

\subsection{Steering Analysis: Focusing on the Feature without Exceptions}
\label{app:vehicle_steering}

Next, we analyzed the use of prompt steering to explicitly nudge GPT-4o to not consider the distracting pattern-related feature in its CoT reasoning chains. For this series of experiments, we used the original features as shown in Table \ref{tab:vehicle_features}. For the first attempt, at the end of both Direct and CoT prompts shown in Tables \ref{tab:vehicle_prompt} and \ref{tab:vehicle_cot_example}, we append the following prompt:
\begin{quote}
    Focus on the license plate of the vehicle and not the other features.
\end{quote}
We ran 30 independent trials each with a maximum of 15 list iterations alloted. Direct prompting eventually achieved perfect classification in every trial in an average of 4.17 trials. CoT prompting still performed worse, achieving perfect classification in only 11 trials, although with a slightly yet statistically significant fewer 10.47 average trials (8.19, 12.74) compared to the original average 12.5 trials needed $(p < 0.05)$. 

To make the steering even more explicit, we also try appending the following prompt:
\begin{quote}
    When trying to predict, first carefully check if any previously seen vehicles match the same license plate. Do your best to keep track along the way.
\end{quote}
We ran 15 independent trials each with a maximum of 15 list iterations alloted. Direct prompting achieved perfect classification in every single trial in either 2 or 3 iterations, with an average of 2.2. We also observe noticeably improved performance with CoT prompting, attaining perfect classification in 14 of the 15 trials, with an average of 3.6 trials needed. 

In an attempt to achieve equally high performance with direct and CoT prompting, we append the following promopt which is a combination of the above two:
\begin{quote}
    Focus on the license plate of the vehicle and not the other features. When trying to predict, first carefully check if any previously seen vehicles match the same license plate. Do your best to keep track along the way.
\end{quote}
We ran 15 independent trials each with a maximum of 15 list iterations alloted. Both direct and CoT prompting manage to achieve perfect classification on the second iteration in each trial.

These findings suggest that, with appropriately designed steering prompts, chain‐of‐thought reasoning can overcome spurious feature distractions and match the efficiency of direct prompting. In particular, explicitly instructing the model to track and prioritize license plate information effectively eliminates reliance on irrelevant pattern‐related features.

\begin{table}[ht!]
\centering
\caption{Example CoT prompt and GPT-4o output for vehicle classification task.}
\label{tab:vehicle_cot_example}
\resizebox{0.98\textwidth}{!}{%
\begin{tabular}{p{\textwidth}}
\hline 
\textbf{Iteration 0:} \\

[Previous conversation, including questions, outputs, and feedback] \\ 
... \\
\textbf{Question 2: } \\ 
The vehicle description is as follows:\\
License plate: G51XVQ\\
Descriptor: Drives in jungles\\
Transmission: Automatic\\
Seat Cover: Cloth\\
Doors: Two\\ \\
Is this vehicle more likely to be a Class A or Class B vehicle? Let's think step by step and answer with 'A' or 'B'. If you are not sure, feel free to guess and explain your reasoning. \\ \\

\textbf{Model output 2: } \\ 
        Given the new vehicle description: \\
        - The vehicle drives in jungles, which suggests it may be a specialized vehicle with off-road capabilities. \\
        - It has an automatic transmission, which is common in both commercial and personal vehicles. \\
        - The seat cover is cloth, which is more common in personal vehicles but not exclusive to them. \\
        - It has two doors, which could indicate a smaller vehicle, but this is not definitive. \\ 
        
        Based on this description alone, there is no clear indication that the vehicle is designed for heavy commercial use, which would typically be associated with Class A vehicles.\\
        The descriptor of driving in jungles implies a need for maneuverability and possibly a smaller size, which might be more characteristic of a Class B vehicle. \\ 
        
        However, since the previous vehicle with a similar non-commercial descriptor was stated to be a Class A vehicle, despite the expectations, it's possible that the classification being used here does not follow the typical commercial vehicle classification system or there may be additional context that is not provided. \\ 
        
        Given the lack of information and the previous correction stating that the glacier-driving vehicle is a Class A vehicle, it's difficult to make an accurate guess. \\
        However, if we follow the same logic as before, we might assume that this jungle-driving vehicle could also be classified as a Class A vehicle, despite the description not aligning with typical Class A characteristics. \\ 
        Therefore, based on the previous correction, the guess would be: \\ 
        'A' \\ 
        But please note that this is a guess and the actual classification could vary based on the specific criteria used for this classification system. \\ \\

\textbf{Feedback 2: } \\
Your answer is incorrect! The vehicle is in class B. \\ \\
\textbf{Question 3: } ... \\
...\\ \\
\textbf{Iteration 1:} \\
...\\
\hline
\end{tabular}%
}
\end{table}


\section{Logical inconsistency task}
Here, participants were tasked to evaluate whether a set of two statements were logically inconsistent. Statement pairs followed two forms: The first statement was always of the form $A \rightarrow B$, where $\rightarrow$ denotes implication, and the second statement was either of the form $A \wedge \neg B$ or $\neg B$, where $\wedge$ denotes the boolean AND operation, and $\neg$ denotes boolean negation. If the second statement was of the form $A \wedge \neg B$, the pair is inconsistent, whereas if the second statement was of the form $\neg B$, the pair is consistent. \citet{khemlani2012hidden} found that if you ask humans to deliberate specifically as to why $A \wedge \neg B$ was plausible, they would subsequently be less accurate at identifying logical inconsistencies between the statements. 

\subsection{Logic dataset generation}
\label{app:logic_generation}

To construct the dataset for the task, we first assigned claims to $A$ and $B$, and then filled in the template to construct the actual statements. To do the first part, we took statements where $A \rightarrow B$ made logical sense following \citet{khemlani2012hidden}. While the original authors simply hand-constructed 12 pairs of claims, we use a combination of natural language inference (NLI) datasets where pairs of statements are filtered to be of the ``entailment'' condition: MNLI, SNLI, and a synthetic datset generated by prompting GPT-4o using the prompt: 

\begin{quote}
    Generate a list of 100 true statements of the format ``if A then B''. For each statement generate the result in JSON format with separate fields for index, A and~B. 
\end{quote}

To construct the actual statements, we fit $A$ and $B$ into the templates in Table~\ref{tab:logic_prompt_template}.

\begin{table}[ht]
\centering
\caption{Sentence template for logical inconsistency task. }
\label{tab:logic_prompt_template}
\resizebox{0.45\textwidth}{!}{%
\begin{tabular}{l}
\hline
\begin{tabular}[c]{@{}l@{}}
\textbf{Statement 1:} If $[A]$, then it is always the case that $[B]$. \\
\textbf{Statement 2 (conflict):} $[A]$, but it is not the case that $[B]$. \\ 
\textbf{Statement 3 (no conflict):} It is not the case that $[B]$.
\end{tabular} \\ \hline
\end{tabular}
}
\end{table}

In addition, to avoid having entailment pairs where statements are more than one sentence long or contain multiple clauses, we limited the maximum amount of words per claim ($A$ or $B$) to seven. This allowed the sentences in the problem to flow smoothly, while still maintaining a large population of entailment pairs. In total, we conducted experiments on 675 pairs from SNLI, 833 pairs from MNLI, and 100 pairs of claims that were synthetically generated, for a final sum of 1608 pairs of $\{A, B\}$. This corresponded to 3216 questions asked per model, over which we calculated model accuracy. 

\subsection{Prompts}
\label{app:logic_prompts}

We prompted models using one zero-shot prompt and two CoT prompts. 
The prompt in the zero-shot condition was as follows: 

\begin{table}[ht]
\centering
\caption{Example prompt for logical inconsistency task, zero shot. }
\label{tab:logic_prompt}
\resizebox{0.75\textwidth}{!}{%
\begin{tabular}{l}
\hline
\textbf{Prompt:} \\
\begin{tabular}[c]{@{}l@{}}The following are two statements:\\
1. $[$Statement 1$]$\\
2. $[$Statement 2$]$\\ \\
Can both of these statements, as explicitly stated, be true at the same time? Please ONLY \\answer with ``Yes" or ``No".
\end{tabular} \\ \hline
\end{tabular}
}
\end{table}

The two chain-of-thought prompts altered the last line in the prompt to the following two sentences, respectively: 
\begin{itemize}
    \item Can both of these statements, as explicitly stated, be true at the same time? Please reason about your answer and then answer ``Yes" or ``No".

    \item Can both of these statements, as explicitly stated, be true at the same time? Please first explain why statement 2 could be true and then answer ``Yes" or ``No".
\end{itemize}

Here, the first prompt follows the standard ``reason about your answer before answering'' CoT request, whereas the latter is a more specific request aimed at more closely replicating the human study. 

For all models, we use temperature = $0.0$. Max tokens was set to 10 for zero-shot and 1000 for CoT. The remaining hyperparameters were set at their default values: top\_p, top\_k, seed, min\_tokens, etc.
\section{Spatial intuition task}
In this task, participants were given drawings of two drinking glasses, one filled with water and one empty. They were asked to estimate the level of water that the second glass would need to be filled to such that the two glasses, when tilted to a certain degree, would have the water they contain reach the rim of the glass at the same angle~\citep{schwartz1999inferences}. 

To simplify the task for the model, we changed the task from drawing a line (image manipulation) to multiple choice (text output) by marking four separate heights on the side of the empty glass, labeling them $A$ through $D$, and asking the model to select a letter. 

\subsection{Motor simulation task dataset generation}
\label{app:water_generation}
To scale up our dataset, instead of fixing the dimensions of the glass that contains water, we varied the width and height in \{2, 3, 4\} and \{4, 5, 6\} respectively (units are per 100 pixels). Then, following \citet{schwartz1999inferences}, we created scenarios where the width and height of the empty cup was \{wider, less wide, same width\} and \{taller, less tall, same height\}. We also varied the amount of water that was in the original glass between \{$\frac{1}{4}, \frac{1}{2}, \frac{3}{4}$\} of its total height. Altogether, this resulted in $243$ unique combinations of problems compared to the original $9$. 

\begin{figure}
    \centering
    \includegraphics[width=0.6\linewidth]{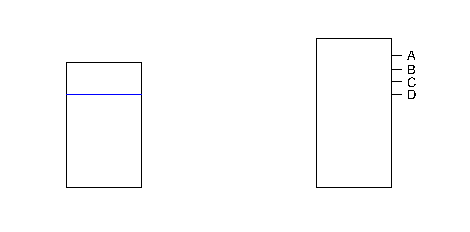}
    \caption{An example of the water problem presented to large multimodal models. The glass on the left is filled with water, and the task is to determine which letter choice the empty glass should be filled to such that when the two glasses tilt to the same angle, water reaches each of their rims at the same time. }
    \label{fig:water_example}
\end{figure}

For each problem, we computed the exact height $h$ that the empty cup would need to be filled with water to in order to get the water to the rim at the desired angle. Then, we sampled from Gaussian noise 
\[
x_i \sim \mathcal{N}(0, \sigma^2)
\]
in order to generate the other answer choices $\{a_i = h+x_i, i\in\{1, 2, 3\}\}$, where $\sigma^2$ is half the distance from the correct answer to the maximum height of the glass. Furthermore, we ensured that none of the answer choices $a_i$ provided were above the maximum height of the cup, below zero, or within distance $\epsilon$ of each other. $\epsilon$ was an empirically determined parameter that controlled the difficulty of the problem, while also having a lower bound due to a limit for how closely the multiple choice letter options could be to each other on the graphical representation of the empty glass. A visual representation of the final problem setup is in Figure~\ref{fig:water_example}.

\subsection{Prompts}
\label{app:water_prompts}
We use one zero-shot prompt and one CoT prompt. The zero-shot prompt is shown in  Table~\ref{tab:water_prompt}. For the CoT prompt, we replaced ``Do not include anything else'' with ``Let's think step by step''. 

For all models, we use temperature = $0.0$. Max tokens was set to 10 for zero-shot and 1000 for CoT. The remaining hyperparameters were set at their default values: top\_p, top\_k, seed, min\_tokens, etc.

\begin{table}[ht]
\centering
\caption{Example prompt for spatial intuition task, zero shot. }
\label{tab:water_prompt}
\resizebox{0.75\textwidth}{!}{%
\begin{tabular}{l}
\hline
\textbf{Prompt:} \\
\begin{tabular}[c]{@{}l@{}}On the left of the figure provided, there is a rectangular glass with its water level indicated \\with the blue line.
On the right, there is a target glass. The target glass has four height \\
markings (A, B, C, D), each indicating a different water level. Which marking should the\\ 
target glass be filled to so that both glasses pour at the same angle?\\\\
Provide your answer using the letter that matches the correct height marking. Do not \\include anything else.\\\\
$[$Prompt image$]$
\end{tabular} \\ \hline
\end{tabular}
}
\end{table}

\section{Working memory preference task}
In this task, participants were shown individual statements about one of four apartments frequently in succession. Each statement describes a different aspect of one apartment, and participants' tasks were to determine which apartment was overall most favorable. However, due to limits in human working memory, their performance in identifying the most beneficial apartment decreased when they tried to reason about the features of each apartment. 

\subsection{Apartment dataset generation}
\label{app:apartment_generation}
To extend this task to LLMs, we scaled up the number of stimuli to hopefully induce an increased pressure on the long-context capabilities of the model. Towards this effort, we first tested the limit of the amount of different features an apartment could have with the help of GPT-4o. We found that the model started repeating aspects of apartments after around 80 unique features. Then, we asked the model to generate positive, negative, and more neutral versions of statements regarding these features:

\begin{itemize}
    \item ``Generate 80 positive statements about different aspects of an apartment. None of the statements should be about the same aspect.''

    \item ``The following are 80 positive statements about aspects of an apartment. For each, generate a corresponding negative statement that is the exact opposite. Make sure that all of the negative statements can coexist with positive statements that are not its direct correspondent. 
    [positive statements]''

    \item ``The following are 80 positive [$\dots$] For each, generate a corresponding neutral statement that is about the same aspect, is worse than the positive version, but is not negative. Make sure that all of the neutral statements can coexist with positive statements that are not its direct correspondent. 
    [positive statements]''

    \item ``The following are 80 negative [$\dots$] For each, generate a corresponding neutral statement that is about the same aspect, is better than the negative version, but is not positive. Make sure that all of the neutral statements can coexist with negative statements that are not its direct correspondent. 
    [negative statements]''

\end{itemize}

We then manually considered conflicts between pairs of statements that were not of the same feature, and manually replaced the only feature statement that had a conflict with another feature. Thus, cohesive descriptions of apartments could be sampled by randomly selecting one of the four statements for each of the 80 features. 

Next, we asked GPT-4o to rate the importance of each statement based on how much the ``statement affects the desirability of the apartment for the average tenant, from -5 to 5, with 5 being most desirable''. Based on this, we could estimate the ground truth quality of each apartment by making the assumption that the features' utilities sum up linearly.\footnote{Note that this is sometimes untrue; e.g., close proximity to a grocery store is much more meaningful when an apartment has a kitchen.} We then randomly sampled apartments with one statement per feature, and computed the score of an apartment as the mean of the feature scores. We then constructed sets of four apartments where the best apartment had at least an average score $\Delta \in \{[0.1, 0.3], [0.3, 0.5], [0.5, 1]\}$ higher than the next-best option. This was to ensure that there is a clear best apartment for the average tenant while not making the task too simple, which were also requirements in the original human study~\citep{dijksterhuis2004think}. Intuitively, $\Delta$ can be considered as a difficulty level, where apartments are closer in rating for lower $\Delta$ problems and are thus harder to get correct. 

Sampling randomly, this led to a total of three datasets corresponding to three ranges of $\Delta$, each containing 100 sets of four apartments. 

Separately, we note that our implementation of this task favors models over humans due to humans being unable to reference the statements after viewing them for the initial 1 second. We recognize that there are other implementations of this task that would be similarly less favorable to models, including simulating partial forgetting by masking some of the sentences. However, since there are no guarantees that performing something like this would be functionally equivalent to how humans process the provided statements, we opted for what we believe is closest to how present models would solve this task in practice.

\subsection{Prompts}
\label{app:apartment_prompts}
For this task, we used one zero-shot and one CoT prompt in our evaluations. The zero-shot prompt is shown in Table~\ref{tab:apartment_prompt}. The CoT prompt replaces ``Respond with only the number of the apartment, do not include anything else.'' with ``Let's think step by step''. 

In our pilot experiments, we also tried a variety of prompts such as replicating the distractor task using a verbal n-back task, setting a time limit for the model (i.e., ``you have three minutes to think about the problem'') or using phrases such as ``very carefully think'' that were present in the original experiment, but the first resulted in too much noise whereas the latter two did not change the results. 

For all models, we use temperature = $0.0$. Max tokens was set to 10 for zero-shot and 8000 for CoT because reasoning chains did not finish in 1000. The remaining hyperparameters were set at their default values: top\_p, top\_k, seed, min\_tokens, etc.

\begin{table}[ht]
\centering
\caption{Example prompt for working memory apartments task, zero shot. }
\label{tab:apartment_prompt}
\resizebox{0.65\textwidth}{!}{%
\begin{tabular}{l}
\hline
\textbf{Prompt:} \\
\begin{tabular}[c]{@{}l@{}}You are an AI assistant designed to evaluate the desirability of four apartments \\
for a potential tenant. You will be given a list of statements about the apartment \\
candidates and how much the tenant likes or dislikes an apartment with the \\ 
quality described by the statement. Your task is to determine which apartment \\
is the most desirable based on the given criteria. \\
The statements are as follows: \\
$[$statements$]$\\\\
Which apartment is most desirable to the tenant? Respond with only the number \\of the apartment, do not include anything else.
\end{tabular} \\ \hline
\end{tabular}
}
\end{table}

\section{Details on Bootstrapping Analyses for the Effectiveness of our Heuristic}
\label{app:effectiveness}

We conduct two analyses to examine the effectiveness of our psychology-inspired heuristic for finding CoT failures: magnitude-based performance decreases and frequency-based decreases irrespective of magnitude. In both bootstrapping analyses, we use the following protocol:

For the larger population, we take all evaluations that compare zero-shot and CoT in a recent metastudy, \citet{sprague2024cot}, for a total of 378. Models evaluated include Llama 2 7b, Mistral 7b, Llama 3.1 8b, Llama 3.1 70b, Gemma 2 9b, Phi-3 Small 8k, Qwen 2 7b, Qwen 2 72b, GPT-4o Mini, Gpt-4o, Claude-3 Haiku, Claude-3.5 Sonnet, Gemini 1.5 Flash, and Gemini 1.5 Pro. Tasks evaluated span various domains such as mathematical reasoning (e.g., GSM8k-Hard), commonsense reasoning (e.g., CommonsenseQA), soft reasoning (e.g., AGIEval LSAT AR Soft Reasoning), and various commonly used benchmarks (e.g., MMLU-Pro, Big-Bench Hard).

For our experiments, we take all comparisons between zero-shot and CoT in our 6 tasks, for a total of 50. These are exactly all of the comparisons that we list in tables 1–6 in the main paper. For task 3, our main metric was number of rounds and not accuracy, so we replaced this with the difference in classification accuracy (e.g., y-axis of Figure~\ref{fig:learning_curve}).

For each comparison, we take the percentage accuracy decrease (consistent with \citet{sprague2024cot}) and use this as the value of the datapoint. We then bootstrap 100,000 samples of size 50 from the population and compute the mean percentage accuracy decrease. None of these 100,000 means were lower than the average percentage mean that we obtained in our experiments.

For each comparison, we labeled accuracy decreases from CoT compared to zero-shot. Separate to the previous analysis, we bootstrapped 100,000 samples of size 50 and counted the number of accuracy decreases. Only 11 of the 100,000 samples had more instances of performance decreases than the 50 datapoints in our experiments.

Both results show that our heuristic is highly statistically significant at finding CoT failure cases.

\end{document}